\documentclass{article}

\usepackage[title]{appendix}%

\usepackage{PRIMEarxiv}
\usepackage{authblk}
\usepackage[utf8]{inputenc} 
\usepackage[T1]{fontenc}    
\usepackage{hyperref}       
\usepackage{multirow}       
\usepackage{url}            
\usepackage{booktabs}       
\usepackage{amsfonts}       
\usepackage{nicefrac}       
\usepackage{microtype}      
\usepackage{lipsum}
\usepackage{fancyhdr}       
\usepackage{graphicx}       
\usepackage{comment}        

\usepackage{array}
\usepackage{float}
\usepackage{subcaption}
\usepackage{siunitx}
\usepackage{algorithm}
\usepackage{algorithmic}
\usepackage{amsmath}
\usepackage{amssymb}
\usepackage{enumitem}
\usepackage{mathtools}
\usepackage{tcolorbox}

\usepackage{rotating}
\usepackage{multirow}
\usepackage{todonotes}

\title{Towards Data-Efficient Cross-Device Generalization of Grad-Shafranov Equilibria via Transfer Learning Neural Operator
}

\author[1]{Jay Phil Yoo}
\author[1]{William Howes}
\author[2]{Yashika Ghai}
\author[1,3]{Kazuma Kobayashi}
\author[1,4,5]{Souvik Chakraborty}
\author[1,3]{Syed Bahauddin Alam}

\affil[1]{Grainger College of Engineering, Nuclear, Plasma \& Radiological Engineering Department, University of Illinois Urbana-Champaign, Urbana, IL, USA
}

\affil[2]{Fusion Energy Division, Oak Ridge National Lab, Oak Ridge, TN, USA
}

\affil[3]{National Center for Supercomputing Applications, Urbana, IL, USA
}

\affil[4]{Department of Applied Mechanics, Indian Institute of Technology Delhi, New Delhi, India
}

\affil[5]{Yardi School of Artificial Intelligence, Indian Institute of Technology Delhi}

\begin{document}
\maketitle

\begin{abstract}
Real-time reconstruction of magnetohydrodynamic equilibria is essential for plasma shaping, stability assessment and feedback control in magnetic confinement fusion. 
However, Grad-Shafranov equilibrium calculations remain largely device-specific and iterative, limiting their use in latency-constrained control 
settings. Existing neural approaches can accelerate 
individual equilibrium predictions, but they do not generally provide reusable models across changing plasma 
boundaries or tokamak geometries. Here we show that equilibrium reconstruction can 
be recast as a cross-device operator learning problem. We develop a domain-specific neural operator framework that maps geometry and profile parameters directly to the poloidal flux 
field, replacing repeated solve-on-demand computation with amortized operator inference. Using the analytically tractable 
Solov’ev family as a controlled Grad-Shafranov testbed, we generate equilibria across eight geometrically distinct tokamak-like configurations and 
benchmark five neural operator architectures under four transfer-learning strategies. Single-geometry pretraining gives poor transfer to unseen devices, whereas multi-geometry pretraining enables data-
efficient adaptation. The 
Wavelet Neural Operator gives the strongest cross-geometry performance, reaching mean relative L2 errors below 4\% with 100 labelled target equilibria and below 2\% with full fine-tuning. The predicted magnetic fields satisfy the divergence-free constraint to 
numerical precision, and four architectures achieve millisecond or sub-millisecond inference. These results identify neural operator pretraining as a route towards reusable, real-time equilibrium inference across fusion device configurations.
\end{abstract}


\section{Introduction}

Magnetic confinement fusion relies on the ability to create, sustain and control high-temperature plasmas whose shape, current distribution and stability evolve continuously during operation \cite{freidberg1987ideal, wesson2011tokamaks}. In tokamaks, this control problem is fundamentally tied to knowledge of the magnetohydrodynamic equilibrium: the force balance between plasma pressure, magnetic fields and boundary geometry that determines the poloidal flux surfaces and, consequently, the operating state of the discharge. In the axisymmetric setting, this equilibrium is commonly described by the Grad--Shafranov equation, a nonlinear elliptic equation that provides the standard formulation for tokamak equilibrium analysis \cite{lao1990equilibrium, pitcher1997experimental}. Accurate equilibrium reconstruction underpins plasma shaping, stability assessment, actuator design and feedback control, and has therefore become central to both experimental operation and reactor design. For decades, the fusion community has relied on high-fidelity equilibrium codes such as EFIT \cite{lao1985reconstruction, huang2016implementation}, VMEC \cite{hirshman1983steepest, hirshman1986three, seal2016parvmec} and DESC \cite{dudt2020desc}, together with parallelized variants developed for large-scale computation, to infer or compute equilibria under realistic diagnostic, geometric and operational constraints. These tools remain indispensable, but their iterative and device-dependent nature creates a persistent computational bottleneck: equilibria must be repeatedly reconstructed as plasma profiles, boundary conditions and machine configurations change, while control-relevant decisions must often be made on much shorter time scales. The challenge is therefore not only to accelerate a single Grad--Shafranov calculation, but to develop a reusable representation of equilibrium physics that can adapt across operating conditions and, ultimately, across fusion devices.

Machine-learning approaches have begun to address this latency gap by replacing repeated numerical solution with learned approximations of the equilibrium map. In particular, several studies have successfully applied physics-informed neural networks (PINNs) to the development of Grad--Shafranov solvers \cite{jang2024grad, merlo2023physics, merlo2021proof, bockenhoff2018reconstruction, wang2024neural}. This approach is attractive because it replaces explicit solver implementation with optimization of a neural representation constrained by the governing equation and boundary conditions. However, the standard PINN framework remains difficult to deploy for real-time equilibrium reconstruction because training is usually performed separately for each equilibrium instance or device-specific configuration. The resulting ``inference'' process is therefore not a single forward evaluation, but a full training-to-convergence procedure that may take seconds or minutes, thereby inheriting part of the latency burden of conventional solvers. This limitation is particularly important in experimental settings, where plasma discharges exhibit continuously evolving pressure and current density profiles and may require hundreds to thousands of distinct equilibrium reconstructions. Moreover, the geometric diversity across existing and planned tokamak devices, characterized by different aspect ratios, cross-sectional shapes and poloidal field coil configurations, further compounds the problem. Because the solution structure of the Grad--Shafranov equation is intrinsically coupled to the boundary geometry and operational domain, a model trained for one device or one class of equilibria need not transfer to another. Consequently, even when highly accurate within a prescribed configuration, instance-wise or device-specific neural models do not by themselves provide a reusable route to equilibrium inference across fusion platforms.

This limitation motivates a shift from learning discrete equilibrium solutions to learning the continuous \textit{solution operator} itself. Data-driven neural networks have been explored for rapid Grad--Shafranov equilibrium prediction \cite{joung2020deep, zheng2024real}, but such models typically operate on fixed input-output discretizations and are therefore tied to a prescribed geometry or computational grid. Neural operators offer a different route \cite{kovachki2023neural}. They are designed to approximate mappings $\mathcal{G}$ between infinite-dimensional function spaces, so that the learned object is not a single solution but the operator that maps input functions, geometric descriptors and physical parameters to the full equilibrium field. In the context of the Grad--Shafranov equation, this corresponds to learning the map from plasma profiles and boundary geometry to the poloidal flux function $\psi$. Architectures such as Deep Operator Networks \cite{lu2021learning}, Fourier Neural Operators \cite{li2020fourier} and Wavelet Neural Operators \cite{tripura2022wavelet} have been developed precisely for this purpose. Their defining advantage is \textit{amortized inference}: once $\mathcal{G}$ has been learned offline, a new equilibrium can be generated through a single forward pass rather than through repeated iterative solution or retraining. This capability has been demonstrated across a range of scientific and engineering problems, including fusion-relevant applications such as edge-localized mode modelling \cite{rahman2024sparsified, carey2025neural}, magnetohydrodynamic benchmarks such as the Orszag--Tang vortex \cite{duarte2025spectral}, and broader settings in computational fluid dynamics \cite{li2020fourier} and electromagnetics \cite{zhang2025data}, with reported inference times in the microsecond-to-millisecond regime. More importantly for fusion, because the learned representation is an operator rather than a device-specific solution map, neural operators provide a natural framework for transfer: knowledge acquired from one family of equilibria or one set of device geometries can, in principle, be reused and adapted to another. Nevertheless, realizing this possibility requires more than fast inference on a fixed geometry. It requires learning representations that remain useful under changes in boundary shape, profile parameters and device configuration, a setting in which cross-device transfer of equilibrium operators remains largely unexplored.

Although neural operator-based approaches have reduced inference latency by orders of magnitude relative to traditional solvers, obtaining a deep neural surrogate model still requires substantial training data and computational resources, and is typically performed independently for each geometry or device configuration. This limitation has motivated growing interest in foundation models and large pretrained operator models for scientific computing \cite{mccabe2024multiple, tripura2025neural}. Such models aim to encode reusable knowledge of complex dynamical systems and to adapt to new downstream configurations with limited additional data and computation. Recently, McCabe et al.~\cite{mccabe2025walrus} demonstrated the feasibility of training a large-scale model on diverse datasets encompassing three-dimensional MHD simulations alongside a broad range of PDE benchmarks. However, broad pretraining does not automatically guarantee efficient adaptation to a specific downstream physics task, and effective transfer may still require extensive fine-tuning of very large models. For fusion applications, an additional challenge is that transfer must occur not only across parameter regimes, but also across distinct device geometries, each with its own aspect ratio, cross-sectional shape, boundary representation and operational domain. This work addresses this gap by developing an efficient domain-specific neural operator framework for ideal MHD equilibria within the Solov'ev analytic family. The Solov'ev solutions provide a tractable and analytically exact subfamily of the Grad--Shafranov equation for axisymmetric tokamak configurations with linearized pressure and current profiles \cite{cerfon2010one}; they do not represent the full generality of the pressure-balance equation $\mathbf{J}\times\mathbf{B}=\nabla p$ or the complexity of free-boundary experimental reconstruction. This restriction is deliberate: it provides a controlled setting in which the role of geometry-conditioned transfer can be isolated and quantified. While recent machine-learning models have achieved high accuracy for specific equilibrium reconstruction tasks or individual device configurations \cite{merlo2023physics, bockenhoff2018reconstruction, jang2024grad, thun2025neural}, such approaches typically learn parametric mappings from low-dimensional inputs to solutions and do not directly address generalization across distinct machine geometries. Here we instead aim to learn the geometry-conditioned structure of the Grad--Shafranov operator across diverse tokamak-like configurations, that is, domain-specific transferable knowledge encoded through a neural operator. We construct a dataset of Solov'ev equilibria spanning eight geometrically distinct configurations and benchmark five neural operator architectures under four transfer-learning strategies. Models are pretrained on five source geometries and adapted to three previously unseen target geometries using limited labelled equilibria. This pretrain-and-fine-tune design tests whether the learned representation captures reusable differential-operator structure rather than merely interpolating within a device-specific parameter space \cite{goswami2022deep}. We find that single-geometry pretraining gives poor zero-shot transfer, whereas multi-geometry pretraining enables data-efficient adaptation to unseen configurations, with the Wavelet Neural Operator achieving the strongest cross-geometry performance.

\vspace{-1mm}
\begin{tcolorbox}[colback=gray!10!white, colframe=blue!50!black, title=\textbf{Key contributions include:}, coltitle=white, fonttitle=\bfseries]
\begin{itemize}
\item \textbf{Domain-specific operator learning framework for Solov'ev MHD equilibria.}
We formulate Grad--Shafranov equilibrium solving (within the Solov'ev analytic family)
as a cross-geometry operator regression problem and introduce a
\emph{pre-train and fine-tune} strategy that encodes transferable,
geometry-conditioned knowledge across diverse tokamak
configurations.
This represents a conceptual advance beyond device-specific
surrogate modeling, shifting from learning discrete solution
instances to learning a generalizable solution operator within this
analytically tractable subfamily.

\item \textbf{First controlled multi-architecture, multi-strategy
benchmark for cross-device MHD transfer.}
We provide the first systematic empirical comparison of five
state-of-the-art neural operator architectures (FNO, WNO, MIONet,
NOMAD, Sp$^2$GNO) across four transfer learning strategies under
eight geometrically distinct tokamak configurations, establishing
rigorous baselines for future surrogate model development in fusion
science.

\item \textbf{Data-efficient cross-geometry generalization with
minimal target supervision.}
Multi-task pre-training on five source geometries enables successful
adaptation to three structurally distinct target configurations
(SPHER, KSTAR, and D3D-ITER) using as few as
\textbf{10 labeled samples} ($0.2\%$ of the full training set).
WNO and FNO achieve mean relative $\mathcal{L}_2$ errors of
${\sim}9.7\%$ and ${\sim}11.3\%$ at 10 samples, and below $4\%$ at
100 samples ($2\%$ of training data), providing concrete
quantitative benchmarks for data-efficient cross-device plasma
equilibrium inference.

\item \textbf{Physics-consistency verification via divergence-free condition.}
We verify that the predicted magnetic field satisfies $\nabla \cdot \mathbf{B} \approx 0$
across all architectures and fine-tuning regimes, with mean absolute divergence scores
on the order of $10^{-8}$, consistent with floating-point precision. This provides
evidence that the neural operators implicitly encode physically consistent
field structure beyond mere curve-fitting.

\item \textbf{Amortized inference validated for real-time fusion
control.}
Four of the five evaluated operator architectures deliver
\textbf{15 times to approximately 1124 times inference speed up}
relative to the parallelized numerical Solov'ev baseline
(average solve time: $\SI{48}{\milli\second}$ using all 72 cores of
an ARM-based Grace CPU; see Section~\ref{sec:amortized} for details).
The graph-based Sp$^2$GNO, at $\SI{146.9}{\milli\second}$ per
sample, is approximately 3 times slower than this parallelized baseline,
confirming that its graph construction overhead is the dominant cost.
A comprehensive profiling of per-sample energy consumption and GPU
memory further characterizes the deployment footprint of each
architecture.

\end{itemize}
\end{tcolorbox}

\vspace{-1mm}

\begin{figure}[h!t]
    \centering
    \includegraphics[width=1\linewidth]{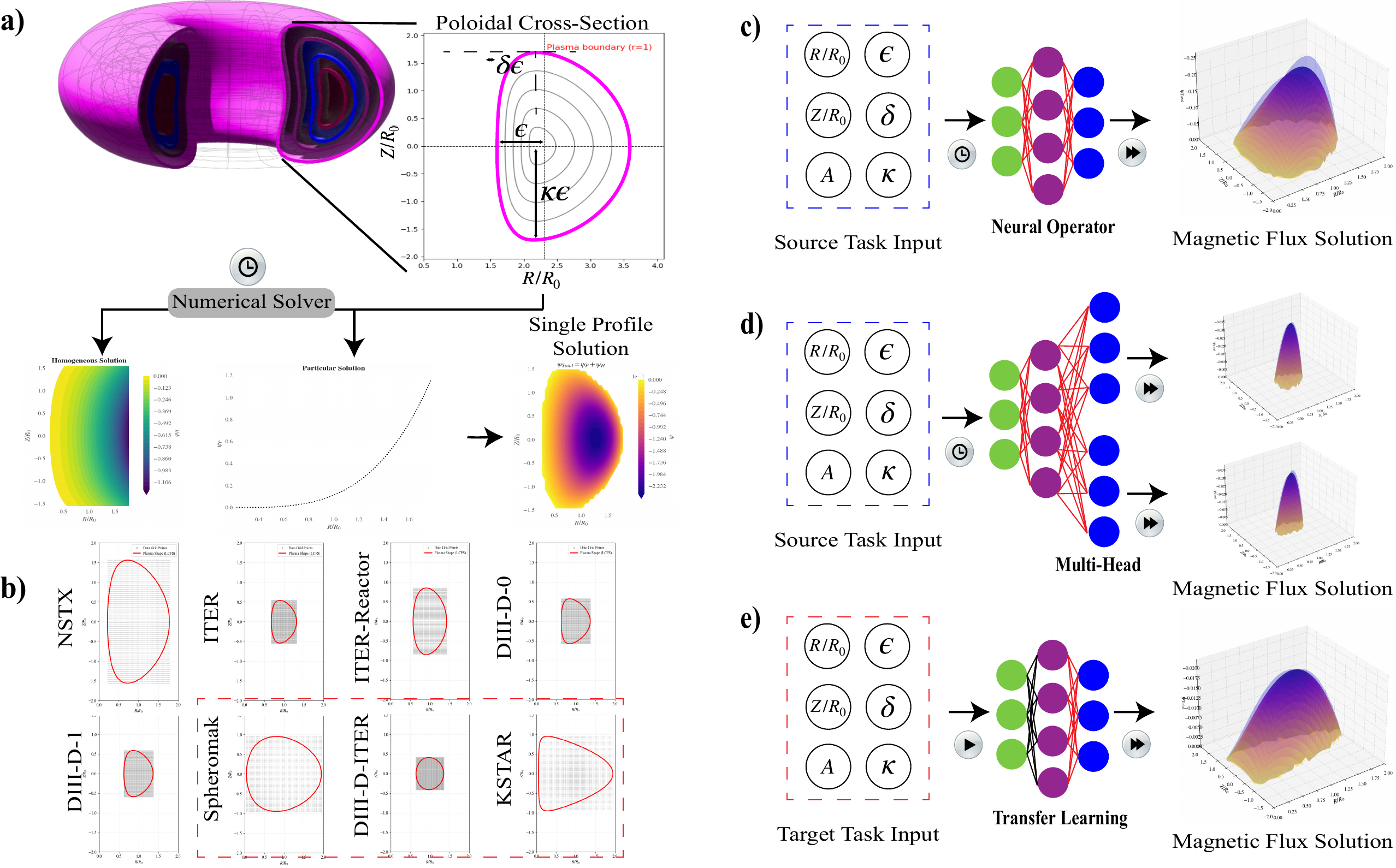}
    \caption{\textbf{Overview of the Neural Operator framework for MHD equilibrium solving and transfer learning across diverse magnetic configurations.} \textbf{(a)} Illustration of a 3D tokamak torus alongside the traditional numerical solver approach, which derives a single profile solution from homogeneous and particular components. The poloidal cross-section defines the plasma boundary via key geometric parameters including inverse aspect ratio ($\epsilon$), elongation ($\kappa$), and triangularity ($\delta$). \textbf{(b)} Representative poloidal geometries for the eight configurations studied; target task geometries are highlighted within red dotted boundaries. All geometries are normalized that R $\in$ [0, 1]. \textbf{(c)} Schematic of the Neural Operator architecture where geometric parameters and the constant $A$ serve as inputs to directly predict the final magnetic flux solution. \textbf{(d)} Visualization of the multi-head configuration where independent parameters are assigned to specific source tasks. \textbf{(e)} The transfer learning mechanism where black connections represent frozen weights from pre-trained models. Fine-tuning is restricted to the final learnable layers (red) using minimal data from unseen target geometries. This approach transitions from computationally intensive pre-training to near-instantaneous inference and accelerated fine-tuning.
    }
\label{fig:main_figure}
\end{figure}

\section{Results}

We first ask whether a neural operator trained on one set of tokamak geometries can learn a representation of Grad--Shafranov equilibrium structure that remains useful when the device geometry changes. This is a substantially harder task than interpolation within a single equilibrium family: the model must not only predict the poloidal flux field for new profile parameters, but also adapt to changes in aspect ratio, elongation, triangularity and boundary shape that alter the spatial structure of the solution manifold. We therefore formulate equilibrium prediction within the Solov'ev family as a geometry-conditioned operator learning problem, $\mathcal{G}: (\boldsymbol{\theta},A)\mapsto\psi$, where $\boldsymbol{\theta}$=($R_0$,$a$,$\varepsilon$,$\kappa$,$\delta$) describes the tokamak boundary geometry and $A$ denotes the profile parameter controlling the analytic equilibrium solution. 
The central objective is not simply to minimize error on geometries seen during training, but to determine whether pretraining across multiple source configurations produces transferable operator representations that can be adapted to previously unseen devices using limited target data.
To test this, we define a controlled source--target transfer protocol. Five geometries, NSTX, ITER, D3D0, D3D1 and ITER-Reactor, are used as source tasks for pretraining, while three geometrically distinct configurations, SPHER, KSTAR and D3D-ITER, are held out as target tasks for adaptation and evaluation. This split allows cross-device generalization to be evaluated explicitly, because no target-geometry samples are used during source pretraining. We benchmark five neural operator architectures, FNO, WNO, MIONet~\cite{jin2022mionet}, NOMAD and Sp$^2$GNO~\cite{sarkar2025spatio}, under four transfer strategies designed to separate device-specific fitting from reusable operator learning. In the \textit{individual} strategy, a model is trained on a single source geometry and directly transferred to the target task. In the \textit{multi-head} strategy, a shared backbone is trained jointly across all five source geometries with independent task-specific projection layers. In the \textit{single-head base} strategy, a unified projection layer is learned across all source tasks and only the final layers are adapted to the target geometry while the pretrained backbone remains frozen. In the \textit{single-head full} strategy, the shared pretrained model is fully fine-tuned on the target geometry. For each target configuration, we evaluate zero-shot prediction and few-shot adaptation using progressively larger labelled target sets drawn from a maximum budget of 1,000 equilibria.

\begin{figure}[h!]
    \centering
    \includegraphics[width=0.75\linewidth]{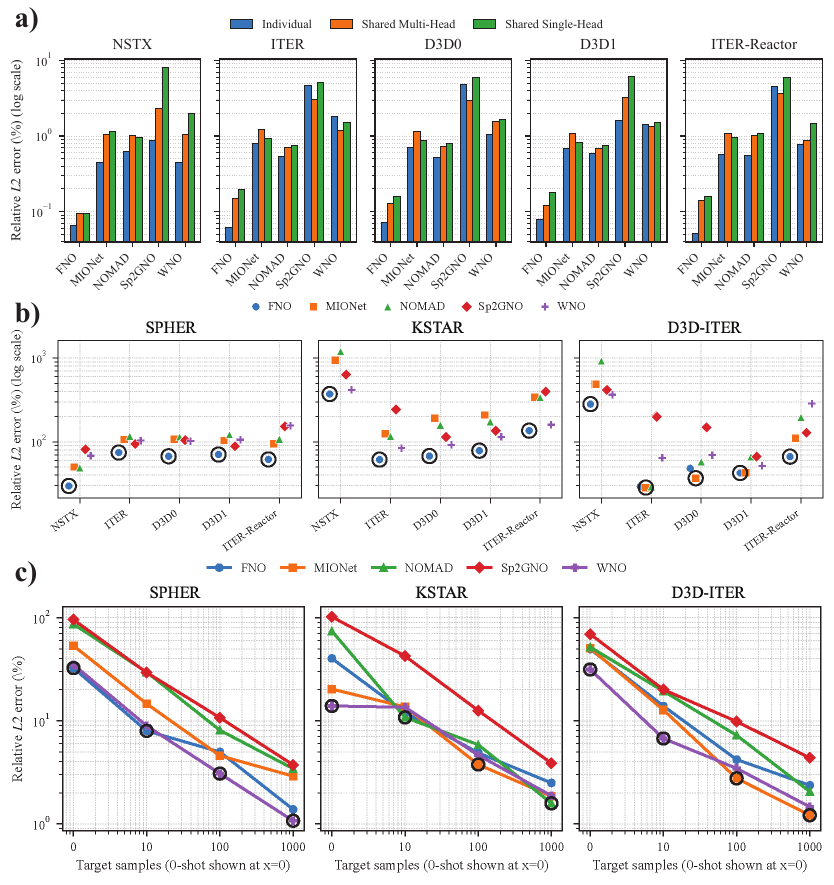}
    \caption{
    \textbf{Full-training and transfer learning performance of neural operators across reactor geometries.}
    All panels report mean relative $L2$ error $\overline{e}_{rel}$ (\%) computed over test examples (Eqs.~\ref{eq:mean_error}) and the relative $L2$ error (Eqs.~\ref{eq:l2}) (lower is better).
    \textbf{(a)} Full-training: end-to-end optimization on each of five source geometries under three training methodologies: \emph{Individual} (separate model per task), \emph{Shared Multi-Head} (shared backbone with task-specific heads), and \emph{Shared Single-Head} (shared backbone with a single head trained jointly on all source tasks).
    \textbf{(b)} Individual transfer (zero-shot) matrix: source-to-target results for all 15 configurations (5 sources $\times$ 3 targets) shown as zero-shot relative $L2$ error for all five operator models; hollow markers denote the best-performing model for each source-to-target configuration.
    \textbf{(c)} Transfer learning summary: performance after transferring from a fully trained source model to each target task under four target set sizes (0, 10, 100, 1000), with the backbone frozen and only the final prediction head fine-tuned.}
    \label{fig:results_figure}
\end{figure}

\subsection{Operator learning on the source geometries}

\begin{table}[h!]
\centering
\caption{Unconstrained Full Training Results with Relative $L2$ Error (\%)}
\begin{tabular}{@{}ccccccc@{}}
\toprule
\textbf{Operator Model} & \textbf{Methodology} &\textbf{NSTX} &\textbf{ITER} &\textbf{D3D0} &\textbf{D3D1} &\textbf{ITER-Reactor}        \\ 
\midrule           
 \multirow{3}{*}{\textbf{FNO}} & Individual & \textbf{0.066} & \textbf{0.062} & \textbf{0.072} & \textbf{0.079} & \textbf{0.051} \\
 & Shared Multi-Head & 0.095 & 0.15 & 0.13 & 0.12 & 0.14 \\
 & Shared Single-Head & 0.094 & 0.20 & 0.16 & 0.018 & 0.16 \\
\cmidrule(l){2-7} 
\multirow{3}{*}{\textbf{MIONet}} & Individual & 0.45 & 0.81 & 0.72 & 0.69 & 0.57 \\
 & Shared Multi-Head & 1.05 & 1.22 & 1.17 & 1.10 & 1.09 \\
 & Shared Single-Head & 1.17 & 0.92 & 0.88 & 0.83 & 0.97 \\
\cmidrule(l){2-7}  
\multirow{3}{*}{\textbf{NOMAD}} & Individual & 0.62 & 0.54 & 0.53 & 0.59 & 0.56 \\
 & Shared Multi-Head & 1.03 & 0.72 & 0.73 & 0.69 & 1.02 \\
 & Shared Single-Head & 0.97 & 0.75 & 0.80 & 0.76 & 1.10 \\
\cmidrule(l){2-7}  
\multirow{3}{*}{\textbf{Sp$^2$GNO}} & Individual & 0.87 & 4.74 & 4.82 & 1.60 & 4.52 \\
 & Shared Multi-Head & 2.34 & 3.06 & 2.99 & 3.30 & 3.72 \\
 & Shared Single-Head & 8.10 & 5.21 & 6.04 & 6.14 & 5.94 \\
 \cmidrule(l){2-7}  
\multirow{3}{*}{\textbf{WNO}} & Individual & 0.45 & 1.83 & 1.06 & 1.45 & 0.78 \\
 & Shared Multi-Head & 1.05 & 1.20 & 1.58 & 1.33 & 0.89 \\
 & Shared Single-Head & 2.00 & 1.51 & 1.67 & 1.50 & 1.47 \\
\bottomrule
\end{tabular}
\label{tab:unconstrained_full}
\end{table}

Before evaluating cross-device transfer, we first assess whether each neural operator can represent the Solov'ev equilibrium map on the source geometries used for pretraining. This source-task experiment provides a capacity check: a model that cannot accurately learn equilibria on the training geometries is unlikely to support meaningful transfer. At the same time, high source-task accuracy is not sufficient evidence of generalization, because a model may fit the solution manifold of each observed geometry while still learning representations that are strongly tied to device-specific boundary structure. We therefore use the source-task results as a baseline against which the subsequent transfer experiments are interpreted.

Table~\ref{tab:combined_unconstrained_full} reports the relative ($L_2$) error for five neural operator architectures trained on the five source geometries, NSTX, ITER, D3D0, D3D1 and ITER-Reactor. Each training sample maps the geometry and profile parameters, $(\boldsymbol{\theta},A)$, to the corresponding equilibrium flux field $\psi$ on a structured two-dimensional computational grid. Across the source tasks, FNO gives the highest in-distribution accuracy, with individual-training errors between (0.062\%) and (0.079\%). This performance is consistent with the smooth, global structure of the Solov'ev flux fields, which can be efficiently represented by Fourier modes. Under shared training, FNO exhibits only a modest loss of accuracy, with errors remaining below (0.20\%), indicating that a common spectral backbone can represent multiple source geometries without severe interference.
MIONet, NOMAD and WNO also learn the source tasks with sub-percent to low-percent errors. NOMAD is the most uniform among these models under individual training, with errors between (0.53\%) and (0.62\%), while MIONet shows a slightly wider range of (0.45\%) to (0.81\%). WNO achieves (0.45\%) error on NSTX and remains within approximately (2\%) across all source geometries and shared-training strategies, although its individual-training error varies more strongly across devices. These results show that several operator families have sufficient expressive capacity to approximate the geometry-dependent Solov'ev solution manifold on observed configurations.
Sp$^2$GNO shows the largest variability across the source tasks. Although it reaches (0.87\%) error on NSTX under individual training, its error increases to (4.74\%) on ITER, (4.82\%) on D3D0 and (4.52\%) on ITER-Reactor. Its performance further deteriorates under shared single-head training, with errors exceeding (5\%) on most source geometries. This suggests that, in the present fixed-grid Solov'ev setting, the graph-spectral representation is more sensitive to geometric heterogeneity and less effective at sharing a unified representation across source domains than the grid-based spectral and operator architectures. The visual representation of the result is presented in Figure~\ref{fig:results_figure}a

These source-task results establish two points that guide the transfer analysis below. First, FNO, WNO, MIONet and NOMAD can accurately learn the equilibrium map on the geometries used for pretraining, so poor transfer cannot be attributed simply to failure of source-task optimization. Second, source-task accuracy alone does not determine cross-device generalization. The next section therefore evaluates whether the learned representations remain useful when the target geometry is unseen during pretraining and only limited target data are available for adaptation.

\begin{figure}[h!]
    \centering
    \includegraphics[width=1\linewidth]{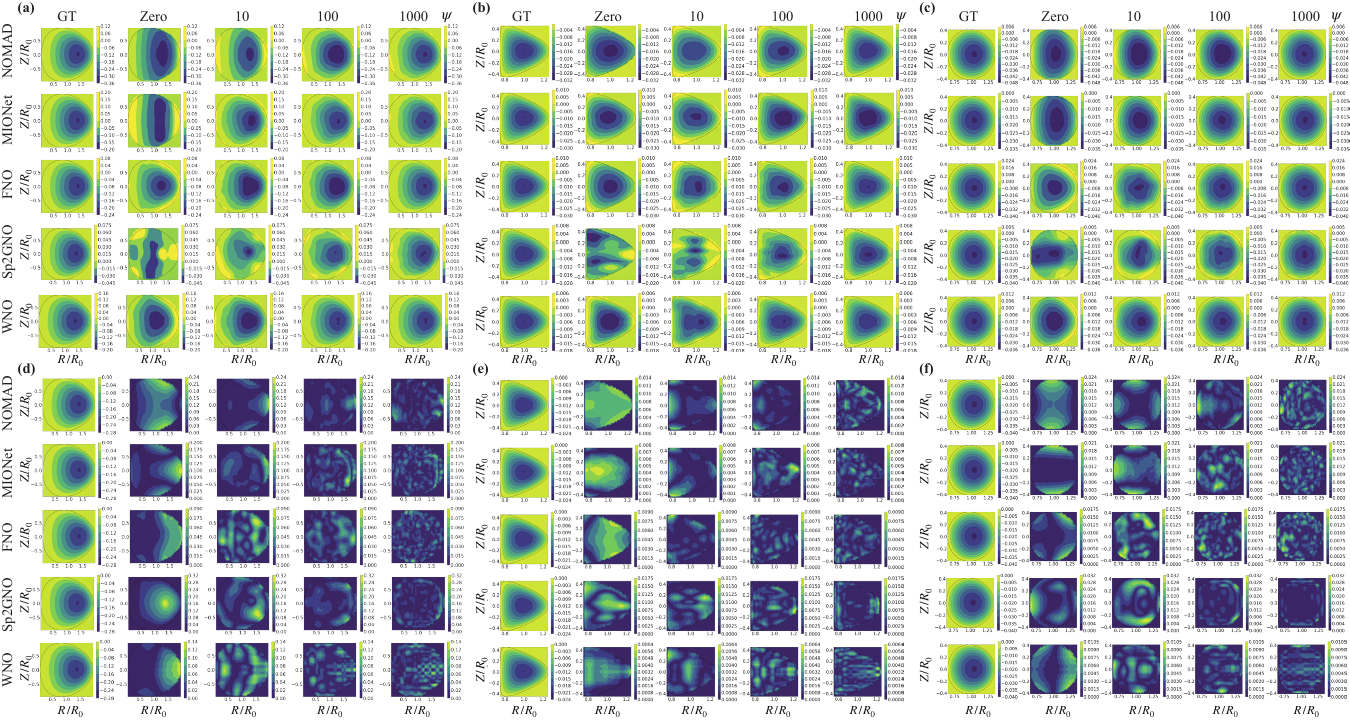}
    \caption{\textbf{Comparative visualization of transfer learning across neural operator architectures for three target configurations:} \textbf{(a)} Spheromak, \textbf{(b)} KSTAR, and \textbf{(c)} D3D-ITER geometries. Magnetic flux profiles ($\Psi$) are shown on the normalized poloidal plane $(R/R_0, Z/R_0)$. Within each panel, rows correspond to different operator models (NOMAD, MIONet, FNO, Sp$^2$GNO, WNO), and columns show the ground truth (GT) followed by predictions obtained in the zero-shot setting and after fine-tuning with 10, 100, and 1,000 target samples. \textbf{(d--f)} Error maps for the corresponding targets, defined as the pointwise Relative L2 deviation $Relative L_2(\Psi_{\mathrm{pred}},\Psi_{\mathrm{GT}})$, highlighting spatial regions where predictions deviate most from the ground truth.}
    \label{fig:comparative_vis}
\end{figure}

\subsection{Generalizability Across Geometries via Transfer Learning}
\begin{table*}[h!]
\centering
\captionsetup{justification=centering}
\caption{Transfer learning results with relative $L2$ error (\%). Left: zero-shot transfer from \emph{individual} source-task pre-training (reported per source task). Right: best transfer from \emph{multi-task} pre-training with recycled single MLP head (zero-shot and few-shot). The top performance for each source-to-target task combination is highlighted in \textbf{bold}.}
\setlength{\tabcolsep}{3.2pt}
\renewcommand{\arraystretch}{1.08}
\begin{tabular}{@{}cc|ccccc|cccc@{}}
\toprule
\multirow{2}{*}{\textbf{Model}} &
\multirow{2}{*}{\textbf{Transfer Task}} &
\multicolumn{5}{c|}{\textbf{Zero Shot from Individual Source Tasks}} &
\multicolumn{4}{c}{\textbf{Best from Multi-Task + Single Head}} \\
\cmidrule(l){3-7}\cmidrule(l){8-11}
& & \textbf{NSTX} & \textbf{ITER} & \textbf{D3D0} & \textbf{D3D1} & \textbf{ITER-R}
& \textbf{ZS} & \textbf{10} & \textbf{100} & \textbf{1000} \\
\midrule
\multirow{3}{*}{\textbf{FNO}} & SPHER  & \textbf{29.72} & \textbf{74.43} & \textbf{67.23} & \textbf{70.52} &\textbf{ 61.72} & \textbf{32.48} & \textbf{7.97} & 4.98 & 1.38 \\
& KSTAR  & \textbf{372.25} & \textbf{61.33} & \textbf{67.63} & \textbf{78.43} & \textbf{136.06} & 40.28 & 11.95 & 4.88 & 2.49 \\
& D3D-ITER & \textbf{282.12} & 29.26 & 48.24 & \textbf{42.50} & \textbf{66.55} & 49.64 & 13.89 & 4.20 & 2.36 \\
\cmidrule(l){2-11}

\multirow{3}{*}{\textbf{MIONet}} & SPHER  & 50.03 & 106.58 & 106.94 & 103.82 & 95.15 & 53.47 & 14.61 & 4.58 & 3.10 \\
& KSTAR  & 944.40 & 124.88 & 191.52 & 209.35 & 339.85 & 20.29 & 17.34 & \textbf{3.83} & 1.95 \\
& D3D-ITER & 485.22 & \textbf{28.43} & \textbf{36.78} & 43.00 & 110.19 & 50.64 & 15.87 & \textbf{3.26} & \textbf{1.25} \\
\cmidrule(l){2-11}

\multirow{3}{*}{\textbf{NOMAD}} & SPHER  & 49.11 & 116.10 & 114.87 & 122.30 & 107.81 & 86.69 & 35.50 & 8.83 & 4.11 \\
& KSTAR  & 1201.73 & 116.47 & 158.75 & 173.58 & 339.60 & 74.36 & \textbf{11.34} & 6.51 & \textbf{1.80} \\
& D3D-ITER & 925.40 & 29.14 & 57.71 & 66.39 & 197.66 & 51.81 & 20.49 & 7.23 & 2.08 \\
\cmidrule(l){2-11}

\multirow{3}{*}{\textbf{Sp$^2$GNO}} & SPHER  & 81.42 & 94.25 & 104.79 & 88.81 & 153.43 & 95.64 & 48.30 & 32.10 & 12.21 \\
& KSTAR  & 634.40 & 243.41 & 114.27 & 135.39 & 397.86 & 102.13 & 48.73 & 19.70 & 10.71 \\
& D3D-ITER & 417.49 & 199.16 & 149.06 & 66.47 & 128.20 & 69.31 & 33.52 & 19.34 & 10.37 \\
\cmidrule(l){2-11}

\multirow{3}{*}{\textbf{WNO}} & SPHER  & 68.31 & 103.59 & 102.14 & 105.97 & 156.06 & 35.05 & 8.92 & \textbf{3.07} & \textbf{1.28} \\
& KSTAR  & 416.95 & 84.30 & 91.92 & 114.63 & 159.75 & \textbf{13.92} & 13.55 & 4.76 & 1.87 \\
& D3D-ITER & 364.64 & 64.43 & 69.40 & 51.58 & 285.61 & \textbf{31.49} & \textbf{6.73} & 3.58 & 1.46 \\
\bottomrule
\end{tabular}
\label{tab:transfer_combined}
\end{table*}

The central test of generalization is each model's ability to adapt to 
unseen target geometries that differ substantially in aspect ratio, 
elongation, and triangularity from the source configurations 
(see Fig.~\ref{fig:main_figure}c). We implement a 
\emph{freeze-then-fine-tune} protocol: the pre-trained weights of each 
operator's core (the Fourier or wavelet layers for spectral 
architectures, and the branch/trunk MLPs for DeepONet-style models) 
are frozen, and only the final projection layers are updated using a 
limited number of labeled target samples. This approach, illustrated 
in Fig.~\ref{fig:main_figure}e, decouples the quality of the 
pre-trained feature representation from the cost of task-specific 
adaptation. We evaluate performance as a function of the fine-tuning 
budget, $N_{\mathrm{target}} \in \{0, 10, 100, 1000\}$, and separate 200 data samples as a test data set. Qualitative 
solution fields and pointwise error maps for all five architectures 
across the three target geometries are provided in 
Fig.~\ref{fig:comparative_vis}.
\begin{figure}[h!]
    \centering
    \includegraphics[width=1.\linewidth]{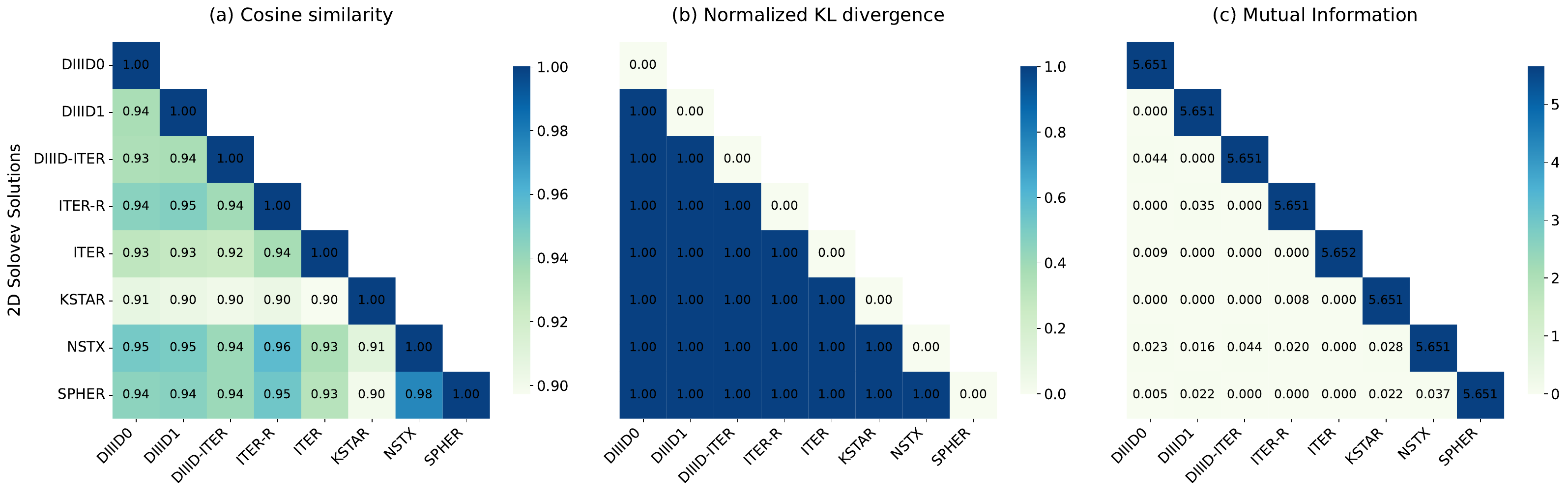}
    \caption{\textbf{Evaluation of 2D Solovev solution similarity across diverse tokamak geometries.} \textbf{(a)} Cosine similarity highlights strong macroscopic structural alignment across all configurations (values $\ge 0.89$). \textbf{(b)} Normalized KL divergence reveals severe spatial domain mismatches, driven by non-overlapping plasma boundaries on the fixed grid. \textbf{(c)} Mutual information demonstrates extremely low point-by-point predictability. Together, these metrics illustrate the inherent difficulty of mapping localized spatial features directly from one reactor geometry to another without adaptation.}
    \label{fig:similarity_metrics}
\end{figure}
\paragraph{Zero-shot transfer from single-task pre-training.}
When each model is pre-trained on a single source geometry and applied directly to a target task without fine-tuning, prediction errors are 
universally prohibitive, as shown in Fig.~\ref{fig:results_figure}(b). 
The best result across all model-source-target combinations is FNO 
pre-trained on NSTX and evaluated on the Spheromak geometry (SPHER), 
with a relative $L_2$ error of $29.72\%$. Averaging the best-source-per-target transfers across all three target tasks for this top-performing architecture already yields a mean error exceeding 
$40\%$. 
This poor zero-shot performance can also be directly anticipated by the similarity metrics established in Fig.~\ref{fig:similarity_metrics}. While the governing MHD physics enforce a broadly consistent macroscopic structure across devices (reflected by high cosine similarities), the precise spatial boundaries differ statistically. The severe spatial domain mismatches (evidenced by the KL divergence) and the lack of point-by-point statistical predictability (near-zero mutual information) mean that an operator trained on the localized spatial distribution of one device cannot directly map to the shifted boundaries of another.

\par
The pattern of zero-shot transfer errors is consistent with the pairwise geometric 
relationships among source and target configurations (Appendix Table~\ref{tab:geometries}). NSTX and the Spheromak both belong to the class of low-aspect-ratio, near-spherical fusion devices, with inverse aspect ratios of $\varepsilon = 0.78$ and $\varepsilon = 0.95$, respectively, and both operate in a regime of weakly elongated, nearly circular cross-sections. This geometric affinity underlies the observation that the model pre-trained on NSTX achieves the lowest zero-shot relative $\mathcal{L}_2$ error on the Spheromak ($29.72\%$ for FNO) among all individual source tasks, while yielding prohibitively large errors on KSTAR ($372.25\%$) and D3D-ITER ($282.12\%$), whose aspect ratios and triangularity values lie far outside the NSTX training distribution. In contrast, models pre-trained individually on ITER achieve the lowest zero-shot errors on the D3D-ITER configuration, with FNO and MIONet reaching $29.26\%$ and $28.43\%$, respectively, consistent with the structural similarity between ITER and D3D-ITER in aspect ratio and elongation. Source geometries D3D0 and D3D1, which share a nearly identical aspect ratio with D3D-ITER ($\varepsilon \approx 0.35$--$0.37$) but differ in triangularity and elongation, yield comparably low but marginally higher zero-shot errors on D3D-ITER relative to ITER-trained models, reflecting partial rather than complete geometric overlap. The ITER-R configuration, representing a reactor-scale device with parameters substantially distinct from the experimental target configurations, consistently produces the largest transfer errors across KSTAR, SPHER, and D3D-ITER among the five source tasks, despite attaining competitive source-task accuracy during individual training. The results confirm that single-task pre-training produces strongly device-specific feature representations that encode the plasma boundary geometry and profile structure of the individual training configuration, rather than the geometry-invariant structure of the underlying Grad--Shafranov operator. Consequently, zero-shot cross-device generalization from a single source geometry is unreliable and fundamentally limited, further motivating the multi-task pre-training framework.

\paragraph{Multi-task pre-training zero-shot and 
few-shot transfer.}
Expanding pre-training to the full five-geometry source suite substantially improves transfer results and reveals a consistent 
architectural hierarchy across fine-tuning regimes 
(Table~\ref{tab:transfer_combined}, right; Fig.~\ref{fig:results_figure}(c)).
\textit{Zero-shot ($N_{\mathrm{target}} = 0$).}
WNO is the strongest architecture in the zero-shot multi-task setting, 
achieving relative $L_2$ errors of $35.05\%$, $13.92\%$, and $31.49\%$ 
on SPHER, KSTAR, and D3D-ITER, respectively (mean: $26.82\%$). It is the 
only architecture to consistently stay below $40\%$ error across all 
three target geometries. MIONet achieves competitive zero-shot 
performance on KSTAR ($20.29\%$) but degrades on the remaining targets 
(mean: $41.47\%$). FNO, NOMAD, and Sp$^2$GNO yield mean zero-shot errors 
of $40.80\%$, $70.95\%$, and $89.03\%$, respectively, showing that 
their learned representations do not transfer reliably to unseen 
boundary geometries without target data.

\textit{Few-shot ($N_{\mathrm{target}} = 10$).}
With only ten labeled fine-tuning samples, all architectures improve 
substantially. WNO and FNO show the strongest gains: WNO achieves a 
mean error of $9.73\%$ ($8.92\%$, $13.55\%$, $6.73\%$ on SPHER, KSTAR, 
and D3D-ITER), while FNO reaches a mean of $11.27\%$, including $7.97\%$ 
on SPHER. MIONet converges to a mean error of $15.94\%$, and NOMAD to 
$22.44\%$. Sp$^2$GNO remains at $43.52\%$, showing limited capacity to 
use the available fine-tuning signal in the extreme low-data regime.

\paragraph{Moderate-to-full fine-tuning.}
At $N_{\mathrm{target}} = 100$ samples 
(Fig.~\ref{fig:results_figure}(c)), FNO, MIONet, and WNO converge to 
a similar performance range with mean relative $L_2$ errors of 
$4.69\%$, $3.89\%$, and $3.80\%$, respectively. NOMAD improves to 
$7.52\%$, while Sp$^2$GNO remains at $23.71\%$, more than five times the 
error of the leading architectures. This gap indicates that the 
graph-based spectral representation in Sp$^2$GNO does not scale 
proportionately with additional target data.
At the full fine-tuning budget of $N_{\mathrm{target}} = 1000$ samples, 
WNO delivers the best overall transfer performance with a mean 
relative $L_2$ error of $1.54\%$, resolving all three target 
geometries below $2\%$ ($1.28\%$ on SPHER, $1.87\%$ on KSTAR, $1.46\%$ 
on D3D-ITER). FNO ($2.08\%$), MIONet ($2.10\%$), and NOMAD ($2.66\%$) 
are all competitive in this regime. Sp$^2$GNO retains a mean error of 
$11.10\%$, failing to converge to acceptable accuracy regardless of 
fine-tuning budget.

\paragraph{Qualitative analysis of predicted solutions and error 
fields.}
The solution fields and error maps in Fig.~\ref{fig:comparative_vis} 
provide qualitative evidence that supports the quantitative findings 
above. MIONet and NOMAD both exhibit a spatial compression artifact in 
the low-data regime ($N_{\mathrm{target}} \leq 10$): the predicted 
magnetic flux field appears geometrically squeezed relative to the 
ground truth, most visibly on SPHER and D3D-ITER but largely absent on 
KSTAR. This artifact diminishes progressively as more target samples 
are added, suggesting that the branch-trunk factorization and 
nonlinear decoder can recover the correct geometry once sufficient 
spatial context is available. FNO and WNO instead produce 
oscillatory, wave-like artifacts near the plasma boundary in the 
few-shot regime; these boundary ripples are more pronounced and 
spatially extensive in FNO than in WNO. Sp$^2$GNO shows the most severe 
failure: the predicted fields cannot recover the correct plasma 
boundary shape even at $N_{\mathrm{target}} = 100$, pointing to a 
fundamental mismatch between its mesh-coupled graph representation 
and the target geometries.

Inspection of the pointwise error fields reveals further distinctions 
across architectures (Fig.~\ref{fig:comparative_vis}(d-f)). FNO, WNO, and MIONet produce spatially structured 
residuals with errors concentrated near high-gradient flux surfaces, a pattern characteristic of local truncation error that diminishes progressively as additional target samples are incorporated during fine-tuning. NOMAD's error fields are the most spatially disorganized across all fine-tuning budgets, consistent with its quantitative underperformance in the transfer setting and suggesting that its nonlinear manifold decoder distributes geometry-specific information broadly across layers rather than localizing it within the final projection, thereby limiting the effectiveness of shallow head adaptation. Sp$^2$GNO exhibits the most severe failure among all architectures: the pointwise error fields retain large-magnitude, indicating that the graph-coupled spectral basis learned during source-task training requires a larger amount of samples to be adjusted to the mesh topology of the target geometries.

\paragraph{Cross-regime analysis: source-task accuracy versus transfer performance.}
An instructive comparison emerges when the source-task training results (Table~\ref{tab:unconstrained_full}) are juxtaposed with the individual zero-shot and multi-task fine-tuning performance (Table~\ref{tab:transfer_combined}, left and right, respectively). The two regimes paint a divergent picture of each architecture's generalization capacity.
FNO achieves the highest source-task fidelity among all architectures, attaining relative $\mathcal{L}_2$ errors below $0.079\%$ under individual training across all five source geometries (Table~\ref{tab:unconstrained_full}); yet applying these individually pre-trained FNO models in a zero-shot setting to unseen target geometries yields errors frequently exceeding $60\%$--$370\%$ (Table~\ref{tab:transfer_combined}, left).
The result that the FNO backbone, which attains sub-$0.1\%$ source-task errors, converges to $1$--$2\%$ transfer errors when only the final projection layers are updated with the full training dataset suggests that the transferable component of the learned equilibrium operator is encoded within the spectral modes of the frozen Fourier convolution blocks.
In contrast, the multi-task pre-training regime (Table~\ref{tab:transfer_combined}, right) demonstrates that as few as $N_{\mathrm{target}} = 10$ labeled samples are sufficient to test the transferable representations consolidated during joint pre-training, e.g., WNO achieves an error of $6.73\%$ on D3D-ITER, and FNO achieves $7.97\%$ error on SPHER.

Sp$^2$GNO presents a contrasting case: the linear projection layers of this architecture are unable to effectively recapture or redirect operator knowledge across geometrically distinct targets without direct modification of the underlying graph formulation, an adaptation that is both conceptually non-trivial.
While MIONet and NOMAD exhibit comparable source-task accuracy (MIONet: $0.45$--$0.81\%$; NOMAD: $0.53$--$0.62\%$ under individual training), NOMAD displays systematically greater difficulty in transferring its learned representations under the fine-tuning protocol. This behavior is consistent with NOMAD's theoretical design.

\paragraph{Architectural interpretation of the transfer hierarchy.}
The performance ordering above can be explained by the inductive 
biases of each architecture. FNO and WNO both represent solution 
features using oscillatory basis functions (global Fourier modes and 
hierarchical wavelet modes, respectively), which are suited to the 
smooth, quasi-periodic character of the Grad-Shafranov equilibrium. 
WNO's multi-resolution wavelet decomposition has a practical advantage 
over FNO in this transfer setting: localized wavelet coefficients 
encode features at multiple spatial scales and can be selectively 
recalibrated when only the final projection layer is updated for a 
new geometry. The global periodicity embedded in FNO's Fourier modes 
is less suited to such targeted recalibration, which accounts for 
WNO's superior performance in the zero-shot and few-shot regimes and 
its lower error at full fine-tuning.
MIONet's branch-trunk factorization separates the functional dependence on the input condition from the spatial evaluation coordinates, preserving transferable physical content within the frozen network while reducing reliance on the outermost layers of each branch and trunk. The linearity of this decomposition allows the fine-tuned projection layers to efficiently recombine pre-trained spatial basis functions for a new target geometry, yielding competitive performance at moderate-to-high fine-tuning budgets.

Sp$^2$GNO encodes solutions by coupling spatial graph topology with 
spectral filtering on the tokamak mesh. While this joint encoding 
produces expressive in-distribution representations, it creates a 
tight dependence between the learned features and the mesh topology 
of the source geometries. When the boundary geometry changes, both 
the effective graph structure and the relevant spectral components 
shift simultaneously. Shallow fine-tuning of the final layers is 
insufficient to bridge this combined shift, which explains the 
persistently high transfer errors of Sp$^2$GNO across all fine-tuning 
budgets.
NOMAD's decoder MLP projects latent embeddings onto task-specific 
solution submanifolds, which drives its competitive in-distribution 
accuracy. However, this nonlinear projection also encodes geometric 
priors from the source distribution in a form that is deeply 
entangled across the decoder's layers and cannot be easily decomposed 
or reused. Updating only the final few projection layers cannot 
re-align these learned submanifolds to a new target geometry, since 
the relevant structural information is distributed rather than localized at the output. This explains NOMAD's performance drop in transfer settings relative to its 
source-task accuracy. Architectures with highly nonlinear, 
geometry-specific decoders may trade transferability for 
in-distribution expressiveness.

\subsection{Physics Consistency: Divergence-Free Verification}
\label{sec:div_free}

A key question for any data-driven surrogate of MHD equilibria is whether the predicted fields are physically consistent beyond reproducing the flux function values. In axisymmetric tokamak geometry, the poloidal magnetic field components derived from the flux function $\psi$ are:
\begin{equation}
    B_R = -\frac{1}{R}\frac{\partial \psi}{\partial Z}, \qquad B_Z = \frac{1}{R}\frac{\partial \psi}{\partial R},
\end{equation}
and the divergence-free condition $\nabla \cdot \mathbf{B} = 0$ in cylindrical coordinates reduces to:
\begin{equation}
    \frac{1}{R}\frac{\partial (R B_R)}{\partial R} + \frac{\partial B_Z}{\partial Z} = \frac{1}{R}\left[\frac{\partial}{\partial R}\left(-\frac{\partial \psi}{\partial Z}\right) + \frac{\partial}{\partial Z}\left(\frac{\partial \psi}{\partial R}\right)\right] = 0,
\end{equation}
which is satisfied identically by any twice-differentiable $\psi$. Although this condition is analytically guaranteed for the exact solution, it is not explicitly enforced during training of the neural operators; its satisfaction therefore provides an independent measure of the physical consistency of the learned flux field and the quality of the gradient estimates from the predicted $\psi$.

We evaluate $\nabla \cdot \mathbf{B}$ numerically via finite differences on the predicted $\psi$ fields for all five architectures across the three target geometries (SPHER, KSTAR, D3D-ITER) and all fine-tuning budgets, using the multi-head transfer learning predictions. The mean absolute divergence scores are reported in Table~\ref{tab:divergence_scores}.
Across all well-trained models and fine-tuning regimes, the divergence scores are on the order of $10^{-8}$, consistent with the expected truncation error of finite-difference gradient estimation at single-precision floating-point ($\sim$$10^{-7}$). This indicates that the neural operator predictions implicitly satisfy $\nabla \cdot \mathbf{B} \approx 0$ to numerical precision, even though this constraint was never explicitly imposed during training. The result is expected for smooth, well-converged flux predictions, as any sufficiently accurate approximation of $\psi$ will inherit the analytical divergence-free property upon numerical differentiation.

\begin{table}[h!]
\centering
\caption{Mean absolute divergence score $|\nabla \cdot \mathbf{B}|$ for multi-head transfer learning predictions across architectures, target geometries, and fine-tuning budgets. Values on the order of $10^{-8}$ are consistent with single-precision finite-difference truncation error, indicating implicit satisfaction of $\nabla \cdot \mathbf{B} \approx 0$.}
\setlength{\tabcolsep}{4pt}
\renewcommand{\arraystretch}{1.05}
\begin{tabular}{@{}lcccccc@{}}
\toprule
\multirow{2}{*}{\textbf{Model}} & \multirow{2}{*}{\textbf{Task}} & \multicolumn{4}{c}{\textbf{Fine-tuning Budget}} \\
\cmidrule(l){3-6}
& & \textbf{0 (ZS)} & \textbf{10} & \textbf{100} & \textbf{Full (1000)} \\
\midrule
\multirow{3}{*}{\textbf{NOMAD}} & SPHER & $7.88\times10^{-8}$ & $8.47\times10^{-8}$ & $8.31\times10^{-8}$ & $8.38\times10^{-8}$ \\
 & KSTAR & $7.24\times10^{-8}$ & $8.39\times10^{-8}$ & $8.21\times10^{-8}$ & $8.30\times10^{-8}$ \\
 & D3D-ITER & $7.60\times10^{-8}$ & $7.88\times10^{-8}$ & $8.17\times10^{-8}$ & $8.67\times10^{-8}$ \\
\cmidrule(l){2-6}
\multirow{3}{*}{\textbf{FNO}} & SPHER & $4.91\times10^{-8}$ & $8.47\times10^{-8}$ & $8.25\times10^{-8}$ & $8.12\times10^{-8}$ \\
 & KSTAR & $1.65\times10^{-7}$ & $8.72\times10^{-8}$ & $8.43\times10^{-8}$ & $8.35\times10^{-8}$ \\
 & D3D-ITER & $1.56\times10^{-7}$ & $8.34\times10^{-8}$ & $9.52\times10^{-8}$ & $8.91\times10^{-8}$ \\
\cmidrule(l){2-6}
\multirow{3}{*}{\textbf{MIONet}} & SPHER & $5.20\times10^{-8}$ & $8.90\times10^{-8}$ & $8.89\times10^{-8}$ & $8.86\times10^{-8}$ \\
 & KSTAR & $7.81\times10^{-8}$ & $9.92\times10^{-8}$ & $9.69\times10^{-8}$ & $9.68\times10^{-8}$ \\
 & D3D-ITER & $5.73\times10^{-8}$ & $8.97\times10^{-8}$ & $9.32\times10^{-8}$ & $9.37\times10^{-8}$ \\
\cmidrule(l){2-6}
\multirow{3}{*}{\textbf{Sp$^2$GNO}} & SPHER & $2.64\times10^{-8}$ & $5.24\times10^{-8}$ & $9.74\times10^{-8}$ & $3.06\times10^{-7}$ \\
 & KSTAR & $1.15\times10^{-8}$ & $1.07\times10^{-8}$ & $3.14\times10^{-9}$ & $5.00\times10^{-9}$ \\
 & D3D-ITER & $2.35\times10^{-8}$ & $1.49\times10^{-8}$ & $1.70\times10^{-8}$ & $3.18\times10^{-8}$ \\
\cmidrule(l){2-6}
\multirow{3}{*}{\textbf{WNO}} & SPHER & $1.14\times10^{-7}$ & $8.38\times10^{-8}$ & $8.02\times10^{-8}$ & $8.10\times10^{-8}$ \\
 & KSTAR & $9.49\times10^{-8}$ & $8.36\times10^{-8}$ & $8.19\times10^{-8}$ & $8.34\times10^{-8}$ \\
 & D3D-ITER & $1.50\times10^{-7}$ & $7.04\times10^{-8}$ & $9.14\times10^{-8}$ & $9.22\times10^{-8}$ \\
\bottomrule
\end{tabular}
\label{tab:divergence_scores}
\end{table}

\subsection{Amortized Inference for Real-Time Application}
\label{sec:amortized}

We verify the real-time inference capability that fundamentally motivates this work by measuring the inference, i.e., solution generation, latency.
Once the one-time offline pre-training is completed (Appendix Table~\ref{tab:model-size-latency-pretrain}),
the iterative computational expense of traditional numerical solvers is replaced by a single,
amortized forward pass through the neural operator.

\paragraph{Numerical solver baseline.}
The Solov'ev analytical solution requires numerically evaluating seven polynomial basis functions and solving a $12\times 12$ linear system for the boundary coefficients at each operating point (see Section~\ref{sec:method}). The baseline solve time reported here was measured on a single compute node equipped with an NVIDIA Grace-Hopper (GH200) Superchip using all 72 cores of its ARM-based Grace CPU and a \texttt{multiprocessing} parallelization framework; no GPU acceleration was applied. Under this fully parallelized configuration, the average solve time per equilibrium sample on a $64\times 64$ grid is $\SI{48}{\milli\second}$. This corresponds to evaluating approximately 4,096 spatial grid points per sample, and the reported latency therefore reflects the cost of generating one complete 2D flux map at the resolution used for training and evaluation. We emphasize that this is a parallelized CPU baseline for an analytic solution; full free-boundary solvers such as EFIT or VMEC operating at experimental resolution and fidelity are substantially more expensive.

\paragraph{Neural operator inference.}
As summarized in Table~\ref{tab:model-size-latency}, inference latencies were measured on a single NVIDIA GH200 GPU using CUDA event timing, which records only the GPU execution time of the forward pass and excludes host-side overhead such as data loading and loss computation. All graph structures and input tensors were transferred to the GPU prior to the timed region so that only the model forward call itself was measured, matching the evaluation methodology used throughout the benchmark. For the four grid-based architectures (FNO, WNO, MIONet, NOMAD), a 10-pass warmup was performed before timing to allow CUDA kernels to reach steady state; per-sample latency was then averaged over the full test dataloader (200 samples at batch size 32). Single-head latency corresponds to the fine-tuned deployment scenario with one projection head; multi-head latency corresponds to evaluating all five task-specific heads in parallel, which is the zero-shot inference setting and thus incurs higher cost.

MIONet achieves the lowest single-head inference latency at $\SI{0.043}{\milli\second}$ per sample, corresponding to a speedup of approximately 1124 times over the numerical baseline. NOMAD and FNO attain single-head latencies of $\SI{0.047}{\milli\second}$ and $\SI{0.073}{\milli\second}$, yielding speedups of approximately 1013 times and 658 times, respectively. WNO exhibits a higher single-head inference latency of $\SI{3.28}{\milli\second}$ per sample, delivering a speedup of approximately 15 times, yet still operating well within the millisecond regime required for real-time control applications. Sp$^2$GNO records a single-head inference latency of $\SI{136.35}{\milli\second}$ per sample, approximately 3 times slower than the parallelized numerical baseline, primarily due to sparse graph message-passing over approximately 51{,}000 edges (KNN with $k{=}50$ on a $32{\times}32$ grid) and per-edge gating MLP evaluation, which inflate measured latency relative to the structured-grid architectures.
These results collectively confirm that the amortized inference paradigm of neural operator learning provides a computationally viable pathway toward real-time magnetohydrodynamic equilibrium reconstruction for four of the five evaluated architectures. The reduction in per-sample inference latency of \textbf{15 times to approximately 1124 times} (for WNO, FNO, NOMAD, and MIONet relative to the parallelized CPU baseline of $\SI{48}{\milli\second}$) validates the proposed pre-train and fine-tune framework as a practical tool for the latency-critical demands of tokamak plasma control systems. Additional evaluations of the GPU memory footprint and power consumption per inference are reported in Appendix Tables~\ref{tab:model-size-latency-pretrain} and~\ref{tab:model_inference_power}, respectively.

\begin{table}[h!]
\centering
\caption{Model complexity and inference latency (ms per sample). All inference latencies are measured on a single NVIDIA GH200 GPU using CUDA event timing over the full 200-sample test set at batch size 32, with a 10-pass GPU warmup prior to timing for neural operator models (FNO, WNO, MIONet, NOMAD). Input tensors and graph structures are moved to GPU before the timed region so that only the model forward pass is measured. \emph{Single Head} corresponds to the fine-tuned deployment scenario (one projection head); \emph{Multi-Head} corresponds to evaluating all five task-specific heads in parallel (zero-shot setting). The numerical baseline (Solov'ev analytic solution, parallelized over 72 CPU cores on GH200) averages $\SI{48}{\milli\second}$ per sample.}
\label{tab:combined-training-time}
\begin{tabular}{@{}ccccccc@{}}
\toprule
\multirow{2}{*}{\textbf{Model}} & \multirow{2}{*}{\textbf{MLP-Head Count}} & \multirow{2}{*}{\textbf{Parameters}} & \multicolumn{3}{c}{\textbf{Training Latency (ms/it)}} & \multirow{2}{*}{\textbf{Inf. Latency (ms/sample)}}    \\ \cmidrule(l){4-6} 
& & & Single Task & Multi-Tasks & Transfer & \\ \midrule
\multirow{2}{*}{\textbf{FNO}}  & Single Head & 2,180,737 & 10.27 & 34.79 & 4.06 & 0.073 \\
 & Multi-Head & 2,247,301 & - & 34.74 & - & 0.090 \\
\cmidrule(l){2-7}
\multirow{2}{*}{\textbf{MIONet}}  & Single Head & 249,088 & 6.50 & 19.70 & 3.58 & 0.043 \\
 & Multi-Head & 579,328 & - & 23.31 & - & 0.051 \\
\cmidrule(l){2-7}
\multirow{2}{*}{\textbf{NOMAD}}  & Single Head & 364,289 & 7.93 & 29.18 & 4.44 & 0.047 \\
 & Multi-Head & 430,953 & - & 27.04 & - & 0.062 \\
\cmidrule(l){2-7}
\multirow{2}{*}{\textbf{Sp$^2$GNO}}  & Single Head & 1,497,053 & 98.59 & 483.16 & 49.56 & 136.35 \\
 & Multi-Head & 1,563,617 & - & 481.88 & - & 147.15 \\
 \cmidrule(l){2-7}
\multirow{2}{*}{\textbf{WNO}}  & Single Head & 1,787,521 & 240.82 & 1,195.53 & 117.47 & 3.2752 \\
 & Multi-Head & 1,854,085 & - & 1,195.04 & - & 3.5037 \\
\bottomrule
\end{tabular}
\label{tab:model-size-latency}
\end{table}

\section*{Limitations}
\label{sec:limitations}

Despite demonstrating data-efficient cross-geometry adaptation for the
Solov'ev family of Grad--Shafranov equilibria, several limitations
constrain the scope of the present study.

\textbf{Restriction to analytic Solov'ev equilibria.}
Our training and evaluation data are generated from the truncated analytic
Solov'ev solutions of Cerfon and Freidberg~\cite{cerfon2010one}. The Solov'ev
family is an analytically exact but physically restricted subset of the
full Grad--Shafranov equation: it assumes axisymmetry, linearized pressure
and current profiles ($p(\psi)$ and $F^2(\psi)$ linear in $\psi$), and
does not represent the complete pressure-balance equation
$\mathbf{J}\times\mathbf{B}=\nabla p$ in its full generality. This choice
enables controlled experimentation across geometries but limits the
diversity of profile physics and solution regularity. In practical
reconstruction and design settings additional constraints arise from diagnostics, free-boundary conditions, and
coil systems as in EFIT/VMEC/DESC-style workflows~\cite{lao1985reconstruction,hirshman1983steepest,dudt2020desc}.
Accordingly, the reported transfer behavior should be interpreted as evidence
that operator backbones can reuse geometry-conditioned features within this
analytic family, rather than as a complete surrogate for experimentally
constrained equilibrium solvers with nonlinear profiles.

\textbf{Domain representation and boundary handling.}
To support grid-based neural operators (FNO/WNO), we evaluate $\psi$ on a fixed
rectangular grid and mask points outside the plasma boundary. This creates a
mismatch between the true computational domain and the model input domain,
which likely contributes to boundary-localized error observed in few-shot
regimes. More faithful treatments could incorporate geometry-aware
parameterizations, boundary-fitted coordinates, signed-distance fields, or
hybrid mesh and grid representations. Relatedly, extending to free-boundary equilibria or mixed
boundary conditions will require additional modeling choices.

\textbf{Limited geometry coverage and extrapolation.}
Although the target set contains geometries that lie outside the convex hull of
the sources in key parameters (notably triangularity for KSTAR and aspect ratio
for the spheromak), the full space of reactor-relevant variations is far larger
than eight configurations. Transfer success can therefore degrade under more
severe extrapolation, different coil topologies, or profile families that alter
the character of the solution operator.

\textbf{Inference cost of graph-based operators.}
Sp$^2$GNO incurs an inference latency of $\SI{146.9}{\milli\second}$ per sample on the A100 GPU, approximately 3 times slower than the parallelized numerical baseline. This overhead arises from the graph construction and eigen-decomposition steps, which do not benefit from the structured-grid parallelism of spectral operators. Addressing this limitation through approximate graph constructions or precomputed graph embeddings remains future work.

\textbf{Scale of neural operators.}
Our models are modest in size relative to recent large-scale foundation-model proposals~\cite{mccabe2025walrus}. This work does not aim to train a general plasma foundation model; we evaluate within the Solov'ev analytic family. Instead, we provide an
empirical comparison of operator architectures and transfer strategies that
suggest which inductive biases facilitate domain-adaptive reuse.

\section*{Conclusion}
\label{sec:conclusion}

Real-time-capable equilibrium inference is a central enabling technology for
robust plasma control in magnetic confinement fusion, yet conventional
equilibrium reconstruction and design pipelines remain dominated by iterative,
latency-limited solvers~\cite{lao1985reconstruction,hirshman1983steepest,dudt2020desc}.
Motivated by the need for amortized inference, we formulated ideal MHD
equilibrium learning---within the Solov'ev analytic family of Grad--Shafranov
solutions, applicable to axisymmetric configurations with linearized profiles---as
an operator regression problem. We evaluated whether neural operators can acquire
transferable, geometry-conditioned structure rather than merely performing
device-specific interpolation, which is a necessary step toward developing
a domain-specific surrogate model capable of operating across diverse tokamak
configurations.

Using a controlled dataset of Solov'ev equilibria~\cite{cerfon2010one} spanning
eight tokamak-like cross-sectional geometries, we pre-trained five distinct operator
architectures on 5,000 samples across five source configurations and assessed transfer 
to three previously unseen geometries under a freeze-then-fine-tune protocol. We additionally
verified that predicted magnetic fields satisfy $\nabla \cdot \mathbf{B} \approx 0$ to
numerical precision across all architectures, providing physics-consistency evidence
for the learned representations. The
results support five main conclusions.

\paragraph{(i) Single-geometry training does not yield reliable cross-device generalization.}
Across all operator classes, zero-shot transfer from a single source geometry
exhibits prohibitive errors. This indicates that strong in-distribution accuracy
does not imply reusable operator representations under geometric shift.

\paragraph{(ii) Multi-geometry pre-training enables few-shot adaptation, with wavelet operators showing the most favorable transfer behavior.}
Joint pre-training across diverse source geometries substantially improves
transfer, and performance improves rapidly with limited target supervision. Specifically, 
fast adaptation was successful with few-shot transfer using only 10 or 100 samples 
(representing 0.2\% and 2\% of the training data size, respectively). The WNO and FNO achieved the best overall 
transfer accuracy. WNO exhibited strong few-shot behavior, managing to produce 
plausible results for all three unseen geometries averaging below 4\% relative $L_2$ error 
with only 100 samples. This is consistent with the hypothesis that spatially localized, 
multi-resolution representations, combining positional and spectral information, are highly 
advantageous for PDEs with axisymmetric assumptions, non-periodic boundaries, and 
geometry-induced feature shifts.

\paragraph{(iii) Graph-based encodings struggle with geometric shifts.}
In contrast to WNO and FNO, graph-based neural operators (e.g., Sp$^2$GNO) did not show 
competitive results when transferring trained knowledge to unseen geometries. The 
geometry-switching process severely hinders the graph encoding, which must currently be 
handled independently. Multi-head structure analyses suggest that for models struggling 
with cross-domain transfer like Sp$^2$GNO, decomposing the problem into 
geometry-specific solvers may be necessary.

\paragraph{(iv) Neural operator predictions are physically consistent.}
All architectures produce flux predictions that implicitly satisfy the
divergence-free condition $\nabla \cdot \mathbf{B} \approx 0$ to numerical
precision ($\sim$$10^{-8}$) across all fine-tuning regimes, despite this
constraint never being explicitly enforced during training. Slightly
elevated divergence scores in poorly-converged regimes (zero-shot for FNO,
full fine-tuning for Sp$^2$GNO on SPHER) correlate with higher $\mathcal{L}_2$
errors, confirming that divergence score serves as a useful proxy for
prediction quality.

\paragraph{(v) Amortized inference provides a practical latency pathway for four of five architectures.}
MIONet, NOMAD, FNO, and WNO deliver 14 times to approximately 940 times
inference-time acceleration relative to the parallelized analytical solver
baseline ($\SI{48}{\milli\second}$ on 72 CPU cores). This confirms that
operator learning can shift the cost of equilibrium computation from online
iteration to offline training, followed by fast forward evaluation and
lightweight adaptation. Sp$^2$GNO is approximately 3 times slower than
the numerical baseline in its current implementation, underscoring the
practical importance of efficient operator backbone design.


Future work should explore more adaptive neural operator structures to overcome current limitations in terms of fusion device configurations. Key directions include: substituting the continuous wavelet 
transform (CWT) used in this work with a discrete wavelet transform (DWT) to achieve 
faster $\mathcal{O}(N)$ computation. For practical applications in fusion, this work can also be explored for extension beyond 2D analytic Solov'ev equilibria to more general GSE solutions with nonlinear profiles and, ultimately, to 3D equilibrium solutions~\cite{hirshman1983steepest, hirshman1986three, dudt2020desc} for complex magnetic field configurations, which are much more computationally challenging. Finally, exploring larger neural operators via 
parallel GPU computations, alongside hardware-software co-design, will be critical for 
further optimizing computational latency and energy consumption.

\section{Methodology}
\label{sec:method}

\subsection{Data}
The MHD equilibrium solutions serve numerous critical applications including diagnostic interpretation, real-time control implementation, and instability identification \cite{lackner1976computation}.
However, parameters relevant to MHD instability analysis cannot be directly measured in MCF devices due to extreme conditions, such as high temperature and pressure. Currently, the equilibrium solving method relies on numerical analysis. In toroidal or cylindrical geometries with a symmetry direction, the Grad-Shafranov equation \cite{grad1958hydromagnetic, shafranov1958magnetohydrodynamical} defines these equilibria, derived from the general ideal-MHD force-balance:
\begin{equation}
    \frac{\partial^2 \psi}{\partial R^2} - \frac{1}{R}\frac{\partial \psi}{\partial R} + \frac{\partial^2 \psi}{\partial Z^2} + \mu_0 R^2 \frac{dp(\psi)}{d\psi} + \frac{1}{2}\frac{dF^2(\psi)}{d\psi} = 0.
    \label{gse}
\end{equation}
 where $\mu_0$ is a vacuum permeability.
Here, the Grad-Shafranov Equation \eqref{gse} describes a 2D MHD equilibrium with a poloidal flux function $\psi(R,Z)$. Since Equation \eqref{gse} is a PDE, a numerical solution exists for this equation given pre-defined pressure profiles $p$ and poloidal current profiles $F$. Furthermore, introducing a current density function,
\begin{equation}
    J_\phi(\psi) = R \frac{dp(\psi)}{d\psi} + \frac{1}{2\mu_0 R} \frac{dF^2(\psi)}{d\psi}
    \label{J_phi}
\end{equation}
simplifies the equation to:
\begin{equation}
\frac{\partial^2 \psi}{\partial R^2}
- \frac{1}{R} \frac{\partial \psi}{\partial R}
+ \frac{\partial^2 \psi}{\partial Z^2} 
= -\mu_0 R J_\phi.
\label{psi_Jphi}
\end{equation}

Solov'ev equilibria introduce a specific choice of $p(\psi)$ and $F^2(\psi)$ that both are linear functions of $\psi$, yielding an analytically exact solution to the GSE under axisymmetric assumptions. While this linearization enables closed-form treatment and controlled benchmarking across device geometries, it restricts the profile physics relative to the full nonlinear GSE.
We employed the method introduced by Cerfon et al., \cite{cerfon2010one}, which introduces arbitrary parameters solving the Solov'ev equilibria: $\mu_0\frac{dp(\psi)}{d\psi} = A$, $\frac{1}{2}\frac{dF^2(\psi)}{d\psi} = C$ where $A$ and $C$ are constants and $A+C = 1$. Normalizing into nondimensional form, $R = R_0 x$, $Z = R_0 y$, $\Psi = \Psi_0 \psi$, where $R_0$ is the major radius of the plasma and $\Psi_0$ is an arbitrary constant, the Grad-Shafranov equation simplifies to:
\begin{equation}
    x \frac{\partial}{\partial x} \left( \frac{1}{x} \frac{\partial \psi}{\partial x} \right) + \frac{\partial^2 \psi}{\partial y^2} = (1 - A) x^2 + A.
    \label{eq:solovev}
\end{equation}

The solution to \eqref{eq:solovev} is of the form $\psi(x,y) = \psi_P(x,y) + \psi_H(x,y)$, where $\psi_P$ is a particular solution and $\psi_H$ is a homogeneous solution, given respectively by:
\begin{equation}
    \psi_p(x, y) = \frac{x^4}{8} + A \left( \frac{1}{2} x^2 \ln x - \frac{x^4}{8} \right).
    \label{eq:psip}
\end{equation}

\begin{equation}
    x \frac{\partial}{\partial x} \left( \frac{1}{x} \frac{\partial \psi_H}{\partial x} \right) + \frac{\partial^2 \psi_H}{\partial y^2} = 0.
    \label{eq:psih}
\end{equation}

A general polynomial solution for this equation, with vertically symmetric plasma, has been derived by Zheng et al. \cite{zheng1996analytical}. This solution is then truncated by Cerfon et al., such that the highest degree polynomials are $R^6$ and $Z^6$.
\begin{equation}
\psi(x, y) = \frac{x^4}{8} + A \left( \frac{1}{2} x^2 \ln x - \frac{x^4}{8} \right) + c_1 \psi_1 + c_2 \psi_2 + c_3 \psi_3 + c_4 \psi_4 + c_5 \psi_5 + c_6 \psi_6 + c_7 \psi_7,
    \label{eq:psi_cefron}
\end{equation}
 and
\begin{align}
    \psi_1 &= 1, \notag \\
    \psi_2 &= x^2, \notag \\
    \psi_3 &= y^2 - x^2 \ln x, \notag \\
    \psi_4 &= x^4 - 4x^2 y^2, \notag \\
    \psi_5 &= 2y^4 - 9y^2 x^2 + 3x^4 \ln x - 12x^2 y^2 \ln x, \notag \\
    \psi_6 &= x^6 - 12x^4 y^2 + 8x^2 y^4, \notag \\
    \psi_7 &= 8y^6 - 140y^4 x^2 + 75y^2 x^4 - 15x^6 \ln x + 180x^4 y^2 \ln x \notag \\
           & \qquad - 120x^2 y^4 \ln x.
    \label{eq:psi_cefron_basis}
\end{align}
The problem thus simplifies to solving for the seven coefficients $c_i$. Cerfon et al. provided an analytical solution to these coefficients utilizing the boundary conditions resulting from the following definition of our geometry axisymmetric boundary shape: 
\begin{equation}
    \begin{array}{c}
x_b=1+\varepsilon \cos (\tau+\arcsin (\delta) \sin (\tau)), \\
y_b=\varepsilon \kappa \sin (\tau),
\end{array}
\label{eq:solovev-boundary}
\end{equation}
where $\varepsilon = \frac{a}{R_0}$ is the inverse aspect ratio between the minor radius ($a$) and the major radius ($R_0$), $\kappa$ is the elongation, $\delta$ is the triangularity and $\tau \in [0, 2\pi)$, as illustrated in Figure \ref{fig:main_figure}a. It is these three geometry parameters ($\varepsilon, \kappa, \delta$) that define the seven coefficients. These geometry parameters are specific to various tokamak reactor designs, and this paper studies the transfer of model ``knowledge'' from a particular group of reactor designs to a set of completely new tokamak geometries with different boundary parameters. In addition, the equation for the particular solution, Eq.~\eqref{eq:psip}, includes the arbitrary constant for the pressure and net poloidal current, $A$. 
As a result, the solution mapping that is presented in this specialized Grad-Shafranov equation becomes:

\begin{equation}
\mathcal{G}:\Omega\times\mathbb{R}^4 \to \mathbb{L}^2(\Omega),
\end{equation}

where $\Omega$ defines the dimensionless 2D coordinate domain $(x,y)$ defined with the boundary in Eq.\ref{eq:solovev-boundary}, $\mathbb{R}^4$ represents the four real scalar quantities that define the solution: $A$, $\varepsilon$, $\kappa$, and $\delta$. The output functional space $\mathbb{L}^2(\Omega)$ represents the full flux solution $\psi(x,y)$ defined over our 2D domain.

In other words, given $A$, the three geometry parameters, and the desired output coordinates, $\mathcal{G}$ defines the operator mapping to the full flux solution $\psi(x,y)$. In this paper, we utilize neural operators and standard backpropagation training algorithms to approximate this mapping, resulting in a modeling framework that does not require retraining for different geometry parameters, coordinates, or constant $A$ values. Ultimately, this ability to quickly calculate the resulting flux without retraining from a given current/pressure can be utilized in finding an MHD equilibrium solution satisfying the ideal MHD equations:
\begin{align}
    J \times B &= \nabla p \\
    \mu_0 J &= \nabla \times B \\
    \nabla \cdot B &= 0
    \label{eq:idealMHD}
\end{align}

To asses the transfer capabilities of the chosen neural operators, we perform full-training on five different reactor configurations with their own geometry parameters. The details to these configurations are shown in Table \ref{tab:geometries}. This paper then transfers learned knowledge from the five tokamak geometries to three new reactors. The pressure and net current constant $A$ is uniformly sampled from the range $[-0.2, 1.0]$. For each device geometry, the coordinate domain is generated to be a unified grid defined by $x\in[1-\varepsilon, 1+\varepsilon]$ and $y\in[-\kappa \varepsilon, \kappa \varepsilon]$. Although the geometry is inconsistent as defined in Eq.\ref{eq:solovev-boundary}, we extend to a regular, unified grid to allow FNO and WNO to provide predictions. The ground truth solution is still dependent on the reactor geometry parameters. Any visual presentation of the model drops evaluation points outside of the boundary.

For each machine configuration, a total of 1200 data samples were generated by uniformly sampling $A \in [-0.2, 1.0]$. From these, 1000 samples were used for training and 200 for testing. Each data sample ($(x,y)$, $A$, $\kappa$, $\varepsilon$, $\delta$, and $\psi(x,y)$) was resolved on a $64 \times 64$ uniform grid within the $R$ and $Z$ bounds defined by the specific geometry where only the coordinates and output flux varies spatially within the grid. Thus, while all datasets share a $64 \times 64$ discrete grid, the underlying physical domains ($R$ and $Z$ coordinates) and, consequently, the spatial resolution (mesh spacing) differ significantly between configurations. For all training and testing, the $64 \times 64$ high-fidelity data was subsampled to a $32 \times 32$ grid (1024 points) to evaluate the models' performance in a sparse, resolution and grid independent setting.
All Solov'ev equilibrium samples were generated on a high-performance compute node equipped with an NVIDIA Grace-Hopper (GH200) Superchip. The generation process was distributed across all 72 cores of the ARM-based Grace CPU using a multiprocessing framework; under this fully parallelized configuration the average numerical solve time per $64\times64$ equilibrium sample is $\SI{48}{\milli\second}$. This solve time, used as the baseline in all speedup comparisons, corresponds to evaluating the analytic Solov'ev solution (seven polynomial basis functions plus a $12\times12$ boundary-coefficient linear system) on the full two-dimensional grid, and does not require GPU acceleration. Neural operator training and inference were performed on a single NVIDIA A100 GPU provided by the Delta cluster from the National Center for Supercomputing Applications (NCSA)~\cite{Delta}.

Our objective is to test whether an operator, trained to learn the geometry-conditioned structure of the ideal MHD equilibrium (Eq.~\eqref{eq:idealMHD}) on a subset of  machine configurations (Source), can be transferred to  different subset of configurations (Target). While the governing GSE (Eq.~\eqref{eq:solovev}) remains the same, the boundary conditions defined by Eq.~\eqref{eq:solovev-boundary} change with the machine's geometric parameters.

\subsection{Similarity Measure}
\label{sec:similarity_scores}

We examined the dissimilarity between the datasets of 2D Solovev equilibrium solutions across different tokamak geometries. This analysis illustrates the inherent geometric and statistical differences among the diverse plasma configurations (e.g., DIII-D, ITER, KSTAR, NSTX, SPHER). To quantitatively measure the dissimilarity between these parametric systems, we map all solutions onto a fixed $64 \times 64$ spatial grid and consider three distinct metrics: (i) cosine similarity (measuring macroscopic geometric alignment), (ii) normalized Kullback-Leibler (KL) divergence (measuring probability distribution divergence), and (iii) mutual information (measuring statistical dependency).

In particular, we compute the pairwise cosine similarity $S_c(\mathcal{U}_i, \mathcal{U}_j)$, pairwise normalized KL divergence $S_{js}(\mathcal{U}_i, \mathcal{U}_j)$, and pairwise mutual information $S_{mi}(\mathcal{U}_i, \mathcal{U}_j)$ between two datasets $\mathcal{U}_i$ and $\mathcal{U}_j$. These datasets contain the 2D PDE solutions for the physical systems $i$ and $j$, where $i, j = 1, \dots, 8$. The metrics are defined using the following formulae:

\begin{equation}
    S_c(\mathcal{U}_i, \mathcal{U}_j) = \frac{\langle \mathcal{U}_i, \mathcal{U}_j \rangle}{\| \mathcal{U}_i \|_F \| \mathcal{U}_j \|_F},
\end{equation}

\begin{equation}
    S_{js}(\mathcal{U}_i, \mathcal{U}_j) = 1 - \exp \left( - \sum_{u} P_i(u \in \mathcal{U}_i) \log \frac{P_i(u \in \mathcal{U}_i)}{P_j(u \in \mathcal{U}_j)} \right),
\end{equation}

\begin{equation}
    S_{mi}(\mathcal{U}_i, \mathcal{U}_j) = \sum_{u,v} P(u \in \mathcal{U}_i, v \in \mathcal{U}_j) \log \frac{P(u \in \mathcal{U}_i, v \in \mathcal{U}_j)}{P(u \in \mathcal{U}_i) P(v \in \mathcal{U}_j)},
\end{equation}

where $\| \cdot \|_F$ denotes the Frobenius norm. The pairwise similarity scores are obtained by averaging the above metrics over $N = 1000$ training samples from each parametric configuration. 

The cosine similarity values demonstrate the extent of macroscopic structural alignment across the geometries. The normalized KL divergence highlights severe spatial domain mismatches; because the plasma boundaries of different reactor designs do not perfectly overlap on the fixed grid, the underlying probability distributions diverge significantly. Finally, the mutual information metric indicates that the datasets share very little common point-by-point information, rendering them almost statistically independent.

\subsection{Neural Operators}
A neural operator approximate nonlinear operators between Hilbert spaces. Unlike standard neural networks that learn mappings between finite-dimensional vectors, $\mathbb{R}^n \to \mathbb{R}^m$, neural operators learn mappings between functions, $\mathcal{G}: \mathcal{F}_1 \to \mathcal{F}_2$. This capability is crucial for solving parametric PDEs, as it allows the network to learn the underlying solution operator itself. Once trained, a neural operator can produce instantaneous solutions for new, unseen input functions (e.g., new boundary conditions or pressure/current quantities) without retraining, a property known as amortized inference.

\paragraph{Multi-Input Operator Network (MIONet)}
The MIONet model \cite{jin2022mionetlearningmultipleinputoperators} is a derivation of the original DeepONet \cite{lu2021learning} architecture, specifically designed to handle multiple sparse input functions/scalars. MIONet utilizes multiple branches where each network process each input function, $u^i(x)$. In practice, $u^i(x)$ is represented by a set of discrete samples (sensors) or a collection of scalar values, and the branch net acts as an encoder to produce a finite-dimensional embedding (a vector of coefficients), $b^i(u^i(x)) \in \mathbb{R}^p$. In parallel, a singular trunk net processes the coordinates of the query location, $\mathbf{y}$, in the output domain. It learns a set of $p$ basis functions, $t(\mathbf{y}) \in \mathbb{R}^p$, that are optimized for the problem. MIONet combines the branch embedding for each input by utilizing a element-wise multiplication scheme across all branches, resulting in the final branch embedding for $k$ different inputs, $b(u^1,...,u^k) = b^1(u^1) \odot ... \odot b^k(u^k)$ 
The final output of the MIONet operator at the query location $\mathbf{y}$ is then computed as the inner product of the final branch embedding and the trunk basis vector. This basis-coefficient viewpoint directly connects MIONets/DeepONets to classical methods like Reduced Order Models (ROMs). The final output is computed as:
\begin{equation}
\mathcal{G}_{\theta}(u^1,...,u^k)(\mathbf{y}) \;=\; \mathcal{S}(b^1(u^1)\odot...\odot b^k(u^k)\odot t(\mathbf{y}))\;=\; \sum_{l=1}^{p} b_l(u^1,...,u^k)\, t_l(\mathbf{y}) \;=\; \mathbf{b}(u^1,...,u^k)^\top \mathbf{t}(\mathbf{y}),
\label{eq:deepONet}
\end{equation}
where $\odot$ is the Hammard element-wise product, $p$ is the latent dimension, $\mathcal{S}$ is a summation over the $p$ elements in the latent embedding, and $\theta$ is the network parameters.

\paragraph{NOMAD}
The MIONet architecture's decoder, which is simply a dot product, are linear with respect to the final branch net's output coefficients. However, in many physical systems, the solutions lie on a low-dimensional nonlinear submanifold. The Nonlinear Manifold Decoder (NOMAD) architecture was introduced to address this specific challenge by employing a fully nonlinear decoder. NOMAD first uses an encoder network, $\mathcal{E}$, to map the multiple input function $u^i(x)$ into a low-dimensional latent representation, $\mathbf{z}^i = \mathcal{E}(u^i) \in \mathbb{R}^m$. The approximation map, $\mathcal{A}$, maps $\mathcal{A} : \mathbb{R}^m \to \mathbb{R}^n$. Lastly, the decoder, $\mathcal{D}$, which is a separate neural network that takes both the approximated vector $\mathcal{A}(\mathbf{z}) \in \mathbb{R}^n \to \mathbf{y}$ to produce the final solution:
\begin{equation}
    \mathcal{G} \approx \mathcal{G}_{\theta} := \mathcal{D} \circ \mathcal{A} \circ \mathcal{E}.
    \label{eq:nomad}
\end{equation}
By allowing for complex, nonlinear interactions between the latent representation $\mathbf{z}$ and the output coordinates $\mathbf{y}$, NOMAD can learn a more flexible and expressive mapping that explicitly parameterizes this nonlinear solution manifold.

\paragraph{Fourier Neural Operators (FNO)}
A different family of neural operators, as proposed in \cite{li2020fourier}, is formulated as an iterative architecture. The input function $\mathbf{u}(x) = [u^1, ...,u^k,\mathbf{y}]$, which is the concatenation of the problem's multiple inputs and evaluation coordinates, is first lifted to a higher-dimensional representation $v_0(x)$, typically with a linear mapping $\mathcal{P}$. This representation is then updated iteratively through a sequence of layers $v_t \to v_{t+1}$ for $t=0, \dots, T-1$.
As defined in the source material, each update is the composition of a non-local integral operator $\mathcal{K}$ and a local, nonlinear activation function $\sigma$:
\begin{equation}
v_{t+1}(x) = \sigma \left( W v_t(x) + \left( \mathcal{K}(u; \phi) v_t \right)(x) \right), \quad \forall x \in D
\label{eq:no_update}
\end{equation}
where $W$ is a local, linear transformation (e.g., a 1x1 convolution) and $\mathcal{K}$ is the non-local integral operator. This operator is defined by a kernel $\kappa$ that can be parameterized by a neural network $\phi$:
\begin{equation}
\left( \mathcal{K}(u; \phi) v_t \right)(x) := \int_D \kappa(x, y; u(x), u(y); \phi) v_t(y) \, dy.
\label{eq:kernel_integral}
\end{equation}
Composing these linear integral operators with non-linear activations allows the network to approximate complex, non-linear operators \cite{li2020fourier}.
By the Convolution Theorem, this computationally expensive operation is simplified to an element-wise multiplication in the Fourier domain. After the final convolution layer, the model is downlifted, with mapping $\mathcal{Q}$, to the final output $\mathcal{G}_{\theta}(u^1,...,u^k)(\mathbf{y}) = \mathcal{Q}(v_T)$.

\paragraph{Wavelet Neural Operators (WNO)}
\cite{tripura2022wavelet} is built on the same kernel integration framework as the FNO, but it uses the Descrete Wavelet Transform (DWT) \cite{selesnick2005dual} instead of the Fourier Transform. Wavelets provide a multi-resolution basis that is localized in both space and frequency. This localization makes WNOs exceptionally well-suited for capturing multi-scale phenomena, sharp gradients, and complex, non-periodic boundaries, which are characteristic of many plasma physics problems.
Analogous to the FNO, the WNO implements the convolution operator by leveraging a convolution theorem in the transform domain. As shown in the source material, the wavelet transform of a convolution can be expressed as the product of the wavelet transforms:
\begin{equation}
    (\mathcal{K}(\phi) * v_t)(x) = W^{-1}(W(\kappa_\phi) \cdot W(v_t))(x),
\end{equation}
where $W$ and $W^{-1}$ are the forward and inverse wavelet transforms, respectively.

The WNO architecture operationalizes this by parameterizing the kernel $\kappa_\phi$ directly in the wavelet space. The non-local update (Eq. \ref{eq:no_update}) is computed by: Transforming $v_t$ to the wavelet domain: $W(v_t)$, performing an element-wise multiplication with a learned tensor $R_{\phi}$. This tensor $R_{\phi}$ is the discrete wavelet representation of the kernel, $W(\kappa_\phi)$, and transforming back to the spatial domain via the inverse wavelet transform ($W^{-1}$). Formally,
\begin{equation}
    (\mathcal{K}(\phi) * v_t)(x) = W^{-1}(R_\phi \cdot W(v_t))(x), \quad \forall x \in D.
\end{equation}

\paragraph{Spatio-Spectral Graph Neural Operator (Sp$^2$GNO)}
\cite{SARKAR2025117659} also builds upon the convolution framework of FNO, but treats the reactor geometry as a graph and utilizes the Graph Fourier Transform (GFT). Evaluation points in the domain $\Omega$ define nodes $\mathbb{V}$ with features $\mathbf{u}(x) = [u^1,...,u^k,\mathbf{y}]$, and a K-Nearest Neighbors algorithm generates the edge set $\mathbb{E}$. Node features are lifted to an intermediate dimension $v_0 \in \mathbb{R}^{d_v}$ via a mapping $\mathcal{P}$, and the convolution in Eq.~\ref{eq:kernel_integral} is approximated through a combined spectral and spatial block.

The spectral block lifts features to the spectral domain using a partial eigendecomposition (via LOBPCG) of the normalized graph Laplacian $\tilde{\mathbf{L}}=\mathbf{I} - \mathbf{D}^{-1/2}\mathbf{A}\mathbf{D}^{1/2}$, yielding the first $m$ eigenmodes $\mathbf{Q}_m \in \mathbb{R}^{n\times m}$. The GFT and its inverse are then $\mathbf{Q}_m^T$ and $\mathbf{Q}_m$, giving:
\begin{equation}
\mathbf{v}_{t+1}^{spectral} = \sigma(\mathbf{Q}_m \cdot\mathbf{K}\times_1\mathbf{Q}^T_m \cdot \mathbf{v}_t + w(\mathbf{v}_t)),
\end{equation}
where $\mathbf{K}$ is a trainable $n\times d_v\times d_v$ tensor, $\sigma$ is a GeLU activation, and $w$ is a linear skip connection.

The spatial block applies a linear layer followed by a gated aggregation of neighbor features, where the gate $\gamma_{uv} \in [0,1]$ is computed from a small neural network over the Lipschitz embeddings of nodes $u,v$ and the edge weight:
\begin{equation}
\mathbf{v}_{t+1}^{spatial}(\mathcal{Y}_n) = \mathbf{\Gamma} \odot \mathbf{A}\mathbf{v}_{t}(\mathcal{Y}_n)\mathbf{W}, \quad \mathbf{\Gamma}=[\gamma_{uv}],
\end{equation}
\begin{equation*}
\gamma_{uv} = \begin{cases}
\sigma_2(\mathbf{W}_3\sigma_1(\mathbf{W}_1[\mathbf{h}_v||\mathbf{h}_u||\mathbf{W}_2 w_{uv}])), & \text{if $u,v$ connected}\\
0, & \text{otherwise}.
\end{cases}
\end{equation*}

A collaboration layer combines both representations, $\mathbf{v}_{t+1} = f([\mathbf{v}_{t+1}^{spatial} || \mathbf{v}_{t+1}^{spectral}])$, allowing joint global and local graph analysis. As in FNO and WNO, the final features $\mathbf{v}_T$ are downlifted via mapping $\mathcal{Q}$ to produce the output.

In this study, we implement and compare five neural operator architectures for learning the mapping from a parametric tokamak boundary description to the corresponding Grad--Shafranov equilibrium flux $\psi$.
Model performance is quantified using the relative $\mathcal{L}_2$ error with respect to the numerical (Solov'ev) reference solution.
Among the evaluated operators, NOMAD and MIONet follow the branch--trunk paradigm and use ReLU activations in their constituent multilayer perceptrons (MLPs).
For the kernel-based operators (FNO, WNO, and Sp$^2$GNO), the lifting map $\mathcal{P}$ is implemented as a linear layer, and the projection/downlifting map $\mathcal{Q}$ is a one-hidden-layer MLP with GeLU activation, where the hidden width equals the input width.
Within the operator blocks, we use architecture-specific nonlinearities: ReLU after FNO blocks, Mish after WNO blocks, and GeLU after the spectral graph block in Sp$^2$GNO.

For branch--trunk operators (NOMAD and MIONet), each scalar conditioning input (the geometric parameter vector $\boldsymbol{\theta}$ and the scalar $A$) is assigned its own branch network, while the spatial coordinates are processed by a separate trunk network.
NOMAD additionally includes a ``combined'' network that concatenates the branch latent vectors with the trunk latent vector and applies an additional MLP to produce the final output.
For the kernel-based operators, the input at each evaluation point is the concatenation of $(\boldsymbol{\theta},A)$ with the local coordinate, enabling the operator to condition on both global geometry parameters and spatial location.
Consequently, FNO and WNO treat the output domain as a structured $2$D grid (i.e., an ``image'') with six input channels, whereas Sp$^2$GNO represents the domain as a graph with six features per node.
The exact architectural specifications used in this work are provided in Table~\ref{tab:model-architectures}.

\subsection{Full-Training and Transfer Learning}
We distinguish between \emph{source} and \emph{target} geometries.
The source geometries (Table~\ref{tab:geometries}) are used to pre-train neural operators, whereas the target geometries are held out and used to evaluate cross-geometry transfer.
Throughout, ``full-training'' denotes end-to-end optimization of all trainable parameters on the source-geometry datasets.

For all architectures, we interpret the final prediction module as a \emph{projection head} (a linear layer or shallow MLP) that maps a latent representation produced by the operator backbone to the flux output.
Transfer learning is implemented by freezing the backbone and adapting only a small set of head parameters for a new target geometry.
Specifically, for WNO/FNO/Sp$^2$GNO the adapted module is the projection/downlifting map $\mathcal{Q}$; for NOMAD we adapt the final two layers of the combined network; for MIONet we adapt the final linear layer in each branch and the trunk.
In all reported experiments, we fine-tune the selected unfrozen head parameters.

\paragraph{Source pre-training strategies.}
We consider three source-training methodologies that differ in how they parameterize and share the projection head across source geometries:

\begin{enumerate}
\item \textbf{Individual-task full-training (single-geometry pre-training).}
A separate model is trained from scratch for each source geometry, yielding five independently optimized parameter sets.
This provides a strong per-geometry baseline but, empirically, models trained on a single geometry can exhibit limited generalization to equilibria from other geometries, despite the shared underlying physics of the governing equation.

\item \textbf{Multi-task pre-training with a \emph{single shared head}.}
A single model is trained jointly on data pooled from all five source geometries, with one projection head shared across all tasks.
This enforces maximal parameter sharing and encourages the backbone and head to represent geometry-conditioned solution families within a unified parameterization.

\item \textbf{Multi-task pre-training with \emph{multiple heads}.}
A single shared backbone is trained jointly across all five source geometries, but each source geometry is assigned a separate task-specific projection head.
This design encourages the backbone to capture geometry-invariant structure while allowing lightweight, geometry-specific readouts.
When transferring to a target geometry, the multi-head module is replaced by a single target head while keeping the backbone frozen.
\end{enumerate}

\paragraph{Target adaptation (transfer) strategies.}
Given a pre-trained backbone, we evaluate two head-initialization strategies for target-geometry transfer:
(1) \textbf{recycled-head fine-tuning}, which initializes the target head from the pre-trained projection head weights (single-head pre-training) and continues optimizing them on target data; and
(2) \textbf{new-head fine-tuning}, which replaces the projection head with a newly instantiated layer and trains it on target data.
Combined with the choice of pre-training regime (individual-task vs.\ multi-task, single-head vs.\ multi-head), this yields a set of transfer configurations that trade off parameter sharing, specialization, and adaptation capacity.

Tables~\ref{tab:transfer_combined} and~\ref{tab:model-size-latency} summarize the resulting computational profiles and transfer-learning performance across these training and adaptation strategies.

For optimization within full-training and finetuning, the relative L2 error is defined:

\begin{equation}
e_{rel}
= \frac{\|\hat{\psi}-\psi\|_{2}}{\|\psi\|_{2}}
= \frac{\sqrt{\sum_{p=1}^{n}\!\left(\hat{\psi}_o(\mathbf{y}_p)-\psi(\mathbf{y}_p)\right)^{2}}}
       {\sqrt{\sum_{p=1}^{n}\!\left(\psi(\mathbf{y}_p)\right)^{2}}},
\label{eq:l2}
\end{equation}

where $\hat{\psi}$ and $\psi$ represents the predicted output and ground truth poloidal flux, and $\mathbf{y}_p$ is the coordinates for evaluation point $p$. To get the reported mean relative value over all examples the following average calculation is performed where $N$ is the number of dataset examples,

\begin{equation}
\label{eq:mean_error}
\overline{e}_{rel}=\frac{1}{N}\sum_{b=1}^{N} e_{rel}.
\end{equation}

Each full-training source task utilized consistent training hyperparameters for each model for fair comparison as well.  For transfer learning, we varied the hyperparameters for better knowledge transfer. Each full training model trained for a maximum of 150 epochs while transfer learning only ran for a maximum of 100 epochs. For learning rate scheduling, a constant decay method was chosen to reduce the gradient step size and improve optimization. To avoid overfitting, early stopping based on validation was implemented to end training when progress was not shown after a certain number of epochs. The complete parameter list utilized for training is shown in Table \ref{tab:training_hyperparameters}.

Lastly, data generation was performed utilizing the analytical solution in Cerfon et al. and imported python libraries for the eight unique geometries in Table \ref{tab:geometries}. Utilizing a singular Nvidia H200/A100 GPU, our operator models were trained on the flux datasets. All latency and energy calculations were performed on an A100 GPU and the \textbf{pynvml} library was utilized for real-time power statistics during training and inference. The hardware was provided by the Delta cluster from the National Center for Supercomputing Applications (NCSA) \cite{Delta}.

\section*{Acknowledgments}
We thank Mark Cianciosa for useful discussions on the topic of equilibrium reconstructions.

This work leveraged Delta and DeltaAI advanced computing and data resources, funded by the National Science Foundation (awards OAC <2005572 and OAC <2320345) and the State of Illinois. Delta and DeltaAI are joint initiatives of the University of Illinois Urbana‑Champaign and NCSA.
\bibliographystyle{unsrt}  
\bibliography{references}  

\appendix
\section{Geometry Parameters}
Table~\ref{tab:geometries} lists the geometric parameters for all
eight tokamak configurations used in this study, distinguished by
their role as source (full-training) or target (transfer) geometries.
The three parameters, inverse aspect ratio $\varepsilon = a / R_0$,
elongation $\kappa$, and triangularity $\delta$, jointly define the
plasma boundary shape via Eq.~\ref{eq:solovev-boundary} and determine
the seven coefficients of the Solov'ev flux solution.

The five source geometries span a moderate range of each parameter:
$\varepsilon \in [0.32, 0.78]$, $\kappa \in [1.55, 2.00]$, and
$\delta \in [0.22, 0.43]$. NSTX and ITER Reactor bracket the
elongation range ($\kappa = 2.00$ and $1.98$), while ITER represents
the lowest inverse aspect ratio in the source set ($\varepsilon =
0.32$). The two DIII-D configurations (D3D0 and D3D1)
share identical aspect ratios ($\varepsilon = 0.37$) with small
differences in elongation ($\kappa = 1.55$ vs.\ $1.60$) and
triangularity ($\delta = 0.43$ vs.\ $0.36$), providing the model
with two closely related in-distribution configurations and testing
its ability to resolve fine geometric variation.

The three target geometries are chosen to probe distinct regions of
parameter space that are outside, or at the boundary of, the source
training distribution. KSTAR has the highest triangularity in the
entire dataset at $\delta = 0.76$, substantially exceeding the
maximum source triangularity of $\delta = 0.43$, and the lowest
inverse aspect ratio ($\varepsilon = 0.25$), making it the most
geometrically extreme target in the elongation-triangularity subspace.
D3D-ITER lies within the source range in aspect ratio
($\varepsilon = 0.35$) and elongation ($\kappa = 1.17$), but has
near-zero triangularity ($\delta = 0.08$), well below any source
configuration. The Spheromak presents the most geometrically distinct
target: its inverse aspect ratio of $\varepsilon = 0.95$ approaches
unity, far exceeding the source maximum of $\varepsilon = 0.78$
(NSTX), and its elongation of $\kappa = 1.00$ corresponds to a
circular cross-section, a shape not represented among the source
geometries. Together, the three targets require the pre-trained models
to extrapolate across aspect ratio, triangularity, and elongation
simultaneously, providing a stringent test of cross-geometry
generalization.

\begin{table}[h!]
\centering
\captionsetup{justification=centering}
\caption{Geometry Configurations of Various Tokamak Reactors}
\begin{tabular}{@{}ccccc@{}}
\toprule
\textbf{Model} & \textbf{Transfer/Full} &\textbf{$\varepsilon = \frac{a}{R_0}$} &\textbf{$\kappa$} &\textbf{$\delta$} \\ 
\midrule            
\textbf{NSTX} & Full & 0.78 & 2.00 & 0.35 \\
 \cmidrule(l){2-5}  
\textbf{SPHEROMAK} & Transfer & 0.95 & 1.00 & 0.20 \\
 \cmidrule(l){2-5}  
\textbf{ITER} & Full & 0.32 & 1.70 & 0.33 \\
 \cmidrule(l){2-5}  
\textbf{D3D0} & Full & 0.37 & 1.55 & 0.43 \\
 \cmidrule(l){2-5}  
\textbf{D3D1} & Full & 0.37 & 1.60 & 0.36 \\ \cmidrule(l){2-5}  
\textbf{D3D-ITER} & Transfer & 0.35 & 1.17 & 0.08 \\
 \cmidrule(l){2-5}  
\textbf{ITER REACTOR} & Full & 0.43 & 1.98 & 0.22 \\
 \cmidrule(l){2-5}  
\textbf{KSTAR} & Transfer & 0.25 & 1.81 & 0.76 \\
\bottomrule
\end{tabular}
\label{tab:geometries}
\end{table}

\section{Evaluation Across Model Structures}

Tables~\ref{tab:unconstrained_10_all}, \ref{tab:unconstrained_100_all},
and \ref{tab:unconstrained_1000_all} report transfer learning performance
across all four pre-training and fine-tuning strategies at
$N_{\mathrm{target}} \in \{10, 100, 1000\}$ labeled samples, as
defined in Section~\ref{sec:method}. Briefly, \textbf{Individual}
denotes a model pre-trained separately on each source task and
transferred directly, with results averaged over all five source tasks
using the original FCN head; \textbf{Multi-Head} uses a shared backbone
trained jointly with task-specific output heads; \textbf{Single-Head
Base} fine-tunes the backbone layers of the shared single-head model
on the target geometry; and \textbf{Single-Head Full} fine-tunes only
the final projection head of the shared single-head model, which
corresponds directly to the primary transfer results reported in
Table~\ref{tab:transfer_combined}.
Table~\ref{tab:unconstrained_avg} summarizes performance averaged over
all four fine-tuning strategies at each sample budget, alongside the
zero-shot multi-task baseline.

\paragraph{Effect of multi-task pre-training.}
Across all architectures and sample budgets, multi-task pre-training
strategies (Multi-Head and Single-Head) consistently outperform the
Individual baseline. The advantage is most pronounced at
$N_{\mathrm{target}} = 10$ samples. NOMAD's strategy-averaged mean
error under Individual pre-training is $40.89\%$, which drops to
$23.06\%$ under Multi-Head; MIONet reduces from $25.74\%$ to $14.47\%$
under the same comparison. For FNO and WNO, the gap is smaller
($13.78\%$ vs.\ $11.27\%$ and $16.61\%$ vs.\ $9.73\%$, respectively),
indicating that these architectures retain more transferable
representations even from single-task pre-training. As
$N_{\mathrm{target}}$ increases to $1000$, the gap between Individual
and multi-task strategies narrows for all architectures, with all
strategies converging to comparable error levels.

\paragraph{Multi-Head versus Single-Head strategies.}
For FNO, MIONet, NOMAD, and WNO, Multi-Head and Single-Head strategies
yield similar mean errors at all sample budgets, with Single-Head Full
matching or marginally outperforming Multi-Head for WNO and FNO in
most configurations. At $N_{\mathrm{target}} = 1000$ samples, the
strategy-averaged means for these four architectures are $2.72\%$
(FNO), $2.50\%$ (MIONet), $3.84\%$ (NOMAD), and $1.80\%$ (WNO),
with no single strategy consistently dominating across all target
geometries.

Sp$^2$GNO departs from this pattern. At $N_{\mathrm{target}} = 10$
samples, its Multi-Head mean is $30.63\%$, whereas Single-Head Base
and Single-Head Full reach $47.15\%$ and $43.52\%$, respectively, a
gap of roughly $13$--$17$ percentage points. This separation persists
at larger sample budgets: at $N_{\mathrm{target}} = 100$ samples,
Sp$^2$GNO Multi-Head achieves $11.03\%$ while both Single-Head strategies
remain near $23.8\%$, and at $N_{\mathrm{target}} = 1000$ samples,
Multi-Head reaches $3.99\%$ versus ${\sim}11.10\%$ for Single-Head
configurations. No other architecture exhibits a comparable
strategy-dependent gap of this magnitude at any sample budget.
This behavior indicates that Sp$^2$GNO's joint spatial-spectral graph
encoding produces latent features that are structurally coupled to
task-specific output pathways. Multi-Head fine-tuning preserves this
coupling through separate projection heads per task, whereas a single
shared projection cannot adequately reconcile features conditioned on
different source mesh topologies, yielding systematically higher errors
at all target sample sizes.

\begin{table}[h!]
\centering
\captionsetup{justification=centering}
\caption{Relative $L2$ Error (\%) for Unconstrained Transfer Learning Strategies with 10 Samples}
\begin{tabular}{@{}cccccc@{}}
\toprule
\textbf{Model} & \textbf{Transfer Task} &\textbf{Individual} &\textbf{Multi-Head} &\textbf{Single-Head Base} &\textbf{Single-Head Full}  \\ 
\midrule           
\multirow{3}{*}{\textbf{FNO}} & SPHER & \textbf{12.79} & \textbf{10.91} & \textbf{8.87} & \textbf{7.97} \\
 & KSTAR & \textbf{18.24} & \textbf{11.85} & 14.85 & \textbf{11.95} \\
 & D3D-ITER & \textbf{10.30} & 13.52 & 16.93 & 13.89\\
  \cmidrule(l){2-6}  
  
\multirow{3}{*}{\textbf{MIONet}} & SPHER & 24.24 & 16.69 & 15.27 & 14.61 \\
 & KSTAR & 30.69 & 13.99 & 13.63 & 17.34 \\
 & D3D-ITER & 22.30 & \textbf{12.73} & 14.56 & 15.87\\
 
 \cmidrule(l){2-6}  
\multirow{3}{*}{\textbf{NOMAD}} & SPHER & 45.70 & 34.08 & 29.63 & 35.50 \\
 & KSTAR & 45.29 & 14.12 & \textbf{10.78} & 11.34 \\
 & D3D-ITER & 31.68 & 20.97 & 19.40 & 20.49\\
 
 \cmidrule(l){2-6} 
\multirow{3}{*}{\textbf{Sp$^2$GNO}} & SPHER & 53.27 & 29.38 & 49.45 & 48.30 \\
 & KSTAR & 63.53 & 42.45 & 56.79 & 48.73 \\
 & D3D-ITER & 46.70 & 20.06 & 35.20 & 33.52\\
 
 \cmidrule(l){2-6}  
\multirow{3}{*}{\textbf{WNO}} & SPHER & 14.01 & 10.96 & 10.17 & 8.92 \\
 & KSTAR & 18.57 & \textbf{13.52} & 19.20 & 13.55 \\
 & D3D-ITER & 17.24 & 22.68 & \textbf{10.02} & \textbf{6.73}\\
\bottomrule
\end{tabular}
\label{tab:unconstrained_10_all}
\caption*{\textit{*Note: Individual Error is averaged over all full-training tasks (with original FCN head).}}
\end{table}

\begin{table}[h!]
\centering
\captionsetup{justification=centering}
\caption{Relative $L2$ Error (\%) for Unconstrained Transfer Learning Strategies with 100 Samples}
\begin{tabular}{@{}cccccc@{}}
\toprule
\textbf{Model} & \textbf{Transfer Task} &\textbf{Individual} &\textbf{Multi-Head} &\textbf{Single-Head Base} &\textbf{Single-Head Full}  \\ 
\midrule           
\multirow{3}{*}{\textbf{FNO}} & SPHER & 9.77 & 5.43 & 4.99 & 4.98 \\
 & KSTAR & 9.77 & 5.45 & 4.79 & 4.88\\
 & D3D-ITER & \textbf{4.61} & 4.71 & 4.54 & 4.20\\
 
 \cmidrule(l){2-6}  
\multirow{3}{*}{\textbf{MIONet}} & SPHER & 9.37 & 6.43 & 4.65 & 4.58 \\
 & KSTAR & 8.21 & \textbf{4.17} & \textbf{3.76} & \textbf{3.83} \\
 & D3D-ITER & 6.18 & 4.57 & \textbf{2.76} & \textbf{3.26}\\
 
 \cmidrule(l){2-6}  
\multirow{3}{*}{\textbf{NOMAD}} & SPHER & 25.95 & 11.62 & 8.09 & 8.83 \\
 & KSTAR & 22.22 & 8.54 & 5.85 & 6.51\\
 & D3D-ITER & 23.60 & 11.08 & 7.48 & 7.23\\
 
 \cmidrule(l){2-6} 
\multirow{3}{*}{\textbf{Sp$^2$GNO}} & SPHER & 26.23 & 10.71 & 31.85 & 32.10 \\
 & KSTAR & 25.90 & 12.56 & 19.05 & 19.70 \\
 & D3D-ITER & 24.17 & 9.83 & 20.54 & 19.34\\
 
 \cmidrule(l){2-6}  
\multirow{3}{*}{\textbf{WNO}} & SPHER & \textbf{6.26} & \textbf{3.89} & \textbf{3.36} & \textbf{3.07} \\
 & KSTAR & \textbf{6.89} & 4.64 & 5.56 & 4.76 \\
 & D3D-ITER & 5.90 & \textbf{3.45} & 3.81 & 3.58\\
\bottomrule
\end{tabular}
\label{tab:unconstrained_100_all}
\end{table}

\begin{table}[h!]
\centering
\captionsetup{justification=centering}
\caption{Relative $L2$ Error (\%) for Unconstrained Transfer Learning Strategies with 1000 Samples}
\begin{tabular}{@{}cccccc@{}}
\toprule
\textbf{Model} & \textbf{Transfer Task} &\textbf{Individual} &\textbf{Multi-Head} &\textbf{Single-Head Base} &\textbf{Single-Head Full}  \\ 
\midrule           
\multirow{3}{*}{\textbf{FNO}} & SPHER & 4.87 & 2.10 & 1.49 & 1.38 \\
 & KSTAR & 4.89 & 3.08 & 2.50 & 2.49\\
 & D3D-ITER & 2.72 & \textbf{2.28} & 2.47 & 2.36\\
 
 \cmidrule(l){2-6} 
\multirow{3}{*}{\textbf{MIONet}} & SPHER & 5.04 & 2.90 & 3.10 & 3.10 \\
 & KSTAR & 3.39 & 2.35 & 1.85 & 1.95 \\
 & D3D-ITER & 2.22 & \textbf{1.65} & \textbf{1.21} & \textbf{1.25}\\
 
 \cmidrule(l){2-6}  
\multirow{3}{*}{\textbf{NOMAD}} & SPHER & 7.24 & 4.30 & 3.43 & 4.11 \\
 & KSTAR & 7.27 & \textbf{1.73} & \textbf{1.58} & \textbf{1.80}\\
 & D3D-ITER & 7.58 & 2.89 & 2.05 & 2.08\\
 
 \cmidrule(l){2-6} 
\multirow{3}{*}{\textbf{Sp$^2$GNO}} & SPHER & 10.91 & 3.73 & 12.61 & 12.21 \\
 & KSTAR & 8.77 & 3.87 & 11.05 & 10.71 \\
 & D3D-ITER & 8.88 & 4.37 & 10.74 & 10.37\\
 
 \cmidrule(l){2-6}  
\multirow{3}{*}{\textbf{WNO}} & SPHER & \textbf{1.62} & \textbf{1.52} & \textbf{1.07} & \textbf{1.28} \\
 & KSTAR & \textbf{2.90} & 2.46 & 1.95 & 1.87 \\
 & D3D-ITER & \textbf{2.17} & 1.70 & 1.64 & 1.46\\
\bottomrule
\end{tabular}
\label{tab:unconstrained_1000_all}
\end{table}

\paragraph{Strategy-averaged summary.}
Table~\ref{tab:unconstrained_avg} consolidates results by averaging
relative $L_2$ error over all four strategies at each sample budget.
At $N_{\mathrm{target}} = 10$ samples, FNO leads with a strategy-
averaged mean of $12.67\%$, followed closely by WNO at $13.80\%$ and
MIONet at $17.66\%$. NOMAD ($26.58\%$) and Sp$^2$GNO ($43.95\%$) are
notably weaker at this budget. At $N_{\mathrm{target}} = 100$
samples, WNO, MIONet, and FNO converge to $4.60\%$, $5.15\%$, and
$5.68\%$; NOMAD reaches $12.25\%$ and Sp$^2$GNO $21.00\%$. At
$N_{\mathrm{target}} = 1000$ samples, WNO retains the best
strategy-averaged mean at $1.80\%$, followed by MIONet ($2.50\%$),
FNO ($2.72\%$), NOMAD ($3.84\%$), and Sp$^2$GNO ($9.02\%$).
Architectures with low strategy-to-strategy variance across all sample
budgets (FNO, WNO, MIONet) are more robust to the choice of
fine-tuning configuration, which is a practical consideration when
the optimal strategy is not known prior to deployment on a new device
geometry.

\begin{table}[h!]
\centering
\captionsetup{justification=centering}
\caption{Average (Over 4 Transfer Methods without Zero-Shot) Relative $L2$ Error (\%) for Unconstrained Transfer Learning}
\begin{tabular}{@{}cccccc@{}}
\toprule
\textbf{Model} & \textbf{Transfer Task} & \textbf{Zero Shot} &\textbf{10 Samples} &\textbf{100 Samples} &\textbf{1000 Samples}  \\ 
\midrule           
\multirow{3}{*}{\textbf{FNO}} & SPHER & \textbf{32.48} & \textbf{10.14} & 6.29 & 2.46 \\
 & KSTAR & 40.28 & \textbf{14.22} & 6.22 & 3.24 \\
 & D3D-ITER & 49.64 & \textbf{13.66} & 4.52 & 2.46\\
 
 \cmidrule(l){2-6}  
\multirow{3}{*}{\textbf{MIONet}} & SPHER & 53.47 & 17.70 & 6.26 & 3.54 \\
 & KSTAR & 20.29 & 18.91 & \textbf{4.99} & 2.39 \\
 & D3D-ITER & 50.64 & 16.37 & \textbf{4.19} & \textbf{1.58}\\
 
 \cmidrule(l){2-6}  
\multirow{3}{*}{\textbf{NOMAD}} & SPHER & 86.69 & 36.23 & 13.62 & 4.77 \\
 & KSTAR & 74.36 & 20.38 & 10.78 & 3.10 \\
 & D3D-ITER & 51.81 & 23.14 & 12.35 & 3.65\\
 
 \cmidrule(l){2-6} 
\multirow{3}{*}{\textbf{Sp$^2$GNO}} & SPHER & 95.64 & 45.10 & 25.22 & 9.87 \\
 & KSTAR & 102.13 & 52.88 & 19.30 & 8.60 \\
 & D3D-ITER & 69.31 & 33.87 & 18.47 & 8.59\\
 
 \cmidrule(l){2-6}  
\multirow{3}{*}{\textbf{WNO}} & SPHER & 35.05 & 11.02 & \textbf{4.15} & \textbf{1.37} \\
 & KSTAR & \textbf{13.92} & 16.21 & 5.46 & \textbf{2.30} \\
 & D3D-ITER & \textbf{31.49} & 14.17 & \textbf{4.19} & 1.74\\
\bottomrule
\end{tabular}
\label{tab:unconstrained_avg}
\caption*{\textit{*Note: Best Zero Shot Performance came from Single Head/Multiple Task Training, utilizing the original MLP weights.}}
\end{table}

\section{Additionally Explored Model Structures}

Two targeted architectural modifications are investigated to address
specific limitations identified in the primary benchmark: an
alternative fine-tuning target for NOMAD, and a skip-connection
augmentation for Sp$^2$GNO. In the primary experiments, the adapted
module for NOMAD during transfer is the final two layers of the
combined network (NOMAD-COMB), which concatenates branch and trunk
latent vectors before projecting to the output. Here we instead adapt
the trunk network directly (NOMAD-TRUNK), motivated by the observation
that the trunk processes spatial evaluation coordinates and may encode
more transferable geometric structure than the combined decoder, which
entangles spatial and input-condition information through nonlinear
mixing. For Sp$^2$GNO, the primary results reveal unstable source-task
training and poor single-head transfer across all sample budgets. A
residual skip connection is added to the collaboration layer, providing a direct gradient path that
bypasses the joint spatial-spectral mixing and stabilizing optimization
across both source and target tasks. WNO is included in both tables as
a fixed reference.

\paragraph{Source-task training.}
Table~\ref{tab:combined_unconstrained_full} reports full-training
results for all four variants. Since the trunk fine-tuning modification
only affects the transfer protocol and not the model architecture
itself, NOMAD-COMB and NOMAD-TRUNK are identical during full training
under Individual and Shared Single-Head methodologies. Under Shared
Multi-Head training, NOMAD-TRUNK achieves a source-task mean of
$0.69\%$ across all five geometries, compared to $0.84\%$ for
NOMAD-COMB, suggesting that the trunk's shared spatial representation
benefits from multi-task training more than the combined decoder.

The skip connection produces a more substantial improvement for
Sp$^2$GNO. Under Individual training, the standard Sp$^2$GNO reaches
errors of $4.74\%$, $4.82\%$, and $4.52\%$ on ITER, D3D0, and
ITER Reactor, respectively. Sp$^2$GNO-Skip reduces these to $1.25\%$,
$1.31\%$, and $0.90\%$, bringing its individual-training profile
in line with WNO ($0.45$--$1.83\%$). Under Shared Multi-Head training,
Sp$^2$GNO-Skip achieves a source-task mean of $1.12\%$ versus $3.08\%$
for the baseline, and under Shared Single-Head training the
improvement is similarly pronounced ($1.39$--$2.11\%$ versus
$5.21$--$8.10\%$). These results confirm that the standard
Sp$^2$GNO's degradation under shared training reported in the main
text is primarily a training stability issue attributable to the
absence of residual paths in the collaboration layer, rather than a
fundamental limitation of the spatio-spectral encoding.

\paragraph{NOMAD: trunk fine-tuning versus combined-network fine-tuning.}
Table~\ref{tab:combined_nomad_wno} compares transfer results for
NOMAD-COMB and NOMAD-TRUNK across all four strategies and sample
budgets. The two variants exhibit a clear and opposing strategy
dependence. For NOMAD-COMB, Single-Head Base (SH-B) is consistently
the best or near-best strategy across all targets and sample sizes:
at $N_{\mathrm{target}} = 10$ it achieves a mean of $19.94\%$
(SPHER: $29.63\%$, KSTAR: $10.78\%$, D3D-ITER: $19.40\%$), and at
$N_{\mathrm{target}} = 1000$ it reaches $2.35\%$ (SPHER: $3.43\%$,
KSTAR: $1.58\%$, D3D-ITER: $2.05\%$). In contrast, NOMAD-TRUNK
performs best under Multi-Head (MH) pre-training at every sample
budget: at $N_{\mathrm{target}} = 10$ its MH mean is $8.22\%$
(SPHER: $8.78\%$, KSTAR: $7.14\%$, D3D-ITER: $8.73\%$), reducing to
$2.12\%$ at $N_{\mathrm{target}} = 100$ and $0.97\%$ at
$N_{\mathrm{target}} = 1000$. The $0.97\%$ mean at full budget is
the lowest transfer error recorded across all NOMAD configurations
and is competitive with WNO's best single-head result of $1.54\%$.

However, NOMAD-TRUNK's strong performance is tightly coupled to
Multi-Head pre-training. Its Single-Head strategies degrade
substantially: at $N_{\mathrm{target}} = 1000$, SH-B yields a mean
of $4.20\%$ (SPHER: $10.07\%$, KSTAR: $1.41\%$, D3D-ITER: $1.11\%$)
and SH-F reaches $4.63\%$, considerably worse than NOMAD-COMB's
SH-B result of $2.35\%$ at the same budget. At
$N_{\mathrm{target}} = 10$, the NOMAD-TRUNK Individual mean is
$19.32\%$, which is already competitive with NOMAD-COMB's best
strategy at the same budget ($19.94\%$), but both Single-Head
strategies remain above $17\%$. This asymmetry indicates that
the trunk network in NOMAD retains geometry-specific spatial features
that are compatible with task-specific projection heads from
Multi-Head pre-training, but are poorly aligned with the unified
projection of Single-Head models. When the trunk is fine-tuned
independently of the combined decoder, the target geometry signal
is absorbed effectively only if the pre-trained head already
provides a geometry-separated readout pathway.

\paragraph{Sp$^2$GNO: effect of the skip connection.}
Table~\ref{tab:skipped_sp2_wno} compares transfer results for
Sp$^2$GNO and Sp$^2$GNO-Skip, again with WNO as reference.
The skip connection produces a large and consistent improvement at
moderate-to-high sample budgets under Multi-Head pre-training.
At $N_{\mathrm{target}} = 100$, Sp$^2$GNO-Skip MH achieves a mean
of $3.82\%$ (SPHER: $3.21\%$, KSTAR: $4.81\%$, D3D-ITER: $3.43\%$),
compared to $11.03\%$ for the baseline Sp$^2$GNO MH. At
$N_{\mathrm{target}} = 1000$, Sp$^2$GNO-Skip MH reaches $0.69\%$
(SPHER: $0.45\%$, KSTAR: $0.91\%$, D3D-ITER: $0.72\%$), which is the
lowest transfer mean recorded across all architectures and strategies
at this budget, outperforming WNO's best result of $1.07\%$ on SPHER
and MIONet's best of $1.25\%$ on D3D-ITER. Individual fine-tuning also
improves substantially: Sp$^2$GNO-Skip Individual at
$N_{\mathrm{target}} = 1000$ achieves a mean of $1.03\%$ versus
$9.52\%$ for the baseline.

The improvement at $N_{\mathrm{target}} = 10$ is less uniform.
Sp$^2$GNO-Skip MH reduces the mean from $30.63\%$ to $28.01\%$,
driven by a large gain on SPHER ($16.27\%$ vs.\ $29.38\%$), but
KSTAR worsens slightly ($46.46\%$ vs.\ $42.45\%$). This suggests
that the skip connection lowers the minimum achievable error once
sufficient target data is available, but does not fundamentally
change the model's behavior in the extreme few-shot regime, where
the geometric mismatch between source and target mesh topologies
remains the primary source of error.

Sp$^2$GNO-Skip retains the Multi-Head strategy preference of the
baseline: at every sample budget, MH outperforms Single-Head
strategies by a substantial margin (e.g., $0.69\%$ vs.\ $1.03\%$
Individual and $1.08\%$ SH-B at 1000 samples). This confirms that
the strategy dependence identified in the main text is a structural property of
the spatio-spectral graph encoding that the skip connection does
not remove. The skip connection improves the model's representational
capacity and training stability without altering its reliance on
task-separated output heads for effective cross-geometry transfer.

\begin{table}[h!]
\centering
\captionsetup{justification=centering}
\caption{Unconstrained Full Training Results with Relative $L2$ Error (\%) for NOMAD, Sp$^2$GNO, and WNO Models}
\begin{tabular}{@{}ccccccc@{}}
\toprule
\textbf{Operator Model} & \textbf{Methodology} & \textbf{NSTX} & \textbf{ITER} & \textbf{D3D0} & \textbf{D3D1} & \textbf{ITER REACTOR} \\ \midrule
\multirow{3}{*}{\textbf{NOMAD-COMB}} & Individual & 0.62 & 0.54 & 0.53 & 0.59 & 0.56 \\
 & Shared Multi-Head & 1.03 & 0.72 & 0.73 & 0.69 & 1.02 \\
 & Shared Single-Head & 0.97 & 0.75 & 0.80 & 0.76 & 1.10 \\ \cmidrule(l){2-7} 
\multirow{3}{*}{\textbf{NOMAD-TRUNK}} & Individual & 0.62 & 0.54 & 0.53 & 0.59 & 0.56 \\
 & Shared Multi-Head & 0.82 & 0.64 & 0.71 & 0.62 & 0.67 \\
 & Shared Single-Head & 0.97 & 0.75 & 0.80 & 0.76 & 1.10 \\ \cmidrule(l){2-7} 
\multirow{3}{*}{\textbf{Sp$^2$GNO}} & Individual & 0.87 & 4.74 & 4.82 & 1.60 & 4.52 \\
 & Shared Multi-Head & 2.34 & 3.06 & 2.99 & 3.30 & 3.72 \\
 & Shared Single-Head & 8.10 & 5.21 & 6.04 & 6.14 & 5.94 \\ \cmidrule(l){2-7} 
\multirow{3}{*}{\textbf{Sp$^2$GNO-Skip}} & Individual & 0.59 & 1.25 & 1.31 & 1.35 & 0.90 \\
 & Shared Multi-Head & 1.06 & 1.14 & 1.20 & 1.23 & 0.97 \\
 & Shared Single-Head & 2.11 & 1.61 & 1.59 & 1.67 & 1.39 \\ \cmidrule(l){2-7} 
\multirow{3}{*}{\textbf{WNO}} & Individual & 0.45 & 1.83 & 1.06 & 1.45 & 0.78 \\
 & Shared Multi-Head & 1.05 & 1.20 & 1.58 & 1.33 & 0.89 \\
 & Shared Single-Head & 2.00 & 1.51 & 1.67 & 1.50 & 1.47 \\ \bottomrule
\end{tabular}
\label{tab:combined_unconstrained_full}
\end{table}

\begin{table}[h!]
\centering
\captionsetup{justification=centering}
\caption{Relative $L2$ Error (\%) for Unconstrained Transfer Learning Strategies across 10, 100, and 1000 Samples for NOMAD and WNO Models}
\begin{tabular}{@{}llcccccccccccc@{}}
\toprule
\multirow{2}{*}{\textbf{Model}} & \multirow{2}{*}{\textbf{Task}} & \multicolumn{4}{c}{\textbf{10 Samples}} & \multicolumn{4}{c}{\textbf{100 Samples}} & \multicolumn{4}{c}{\textbf{1000 Samples}} \\ \cmidrule(lr){3-6} \cmidrule(lr){7-10} \cmidrule(lr){11-14}
 &  & \textbf{Ind.} & \textbf{MH} & \textbf{SH-B} & \textbf{SH-F} & \textbf{Ind.} & \textbf{MH} & \textbf{SH-B} & \textbf{SH-F} & \textbf{Ind.} & \textbf{MH} & \textbf{SH-B} & \textbf{SH-F} \\ \midrule
\multirow{3}{*}{\rotatebox[origin=c]{90}{\tiny NOMAD-C}} & SPH & 45.70 & 34.08 & \textbf{29.63} & 35.50 & 25.95 & 11.62 & \textbf{8.09} & 8.83 & 7.24 & 4.30 & \textbf{3.43} & 4.11 \\
 & KST & 45.29 & 14.12 & \textbf{10.78} & 11.34 & 22.22 & 8.54 & \textbf{5.85} & 6.51 & 7.27 & 1.73 & \textbf{1.58} & 1.80 \\
 & D3I & 31.68 & 20.97 & \textbf{19.40} & 20.49 & 23.60 & 11.08 & 7.48 & \textbf{7.23} & 7.58 & 2.89 & \textbf{2.05} & 2.08 \\ \cmidrule(l){2-14} 
\multirow{3}{*}{\rotatebox[origin=c]{90}{\tiny NOMAD-T}} & SPH & 21.59 & \textbf{8.78} & 21.57 & 24.26 & 12.05 & \textbf{2.28} & 13.75 & 13.72 & 9.62 & \textbf{0.96} & 10.07 & 11.44 \\
 & KST & 20.11 & \textbf{7.14} & 11.71 & 14.59 & 4.14 & \textbf{2.33} & 3.29 & 3.96 & 1.50 & \textbf{0.99} & 1.41 & 1.09 \\
 & D3I & 16.26 & 8.73 & \textbf{8.70} & 17.86 & 3.23 & \textbf{1.74} & 2.93 & 5.33 & 1.26 & \textbf{0.95} & 1.11 & 1.35 \\ \cmidrule(l){2-14} 
\multirow{3}{*}{\rotatebox[origin=c]{90}{\tiny WNO}} & SPH & 14.01 & 10.96 & 10.17 & \textbf{8.92} & 6.26 & 3.89 & 3.36 & \textbf{3.07} & 1.62 & 1.52 & \textbf{1.07} & 1.28 \\
 & KST & 18.57 & \textbf{13.52} & 19.20 & 13.55 & 6.89 & \textbf{4.64} & 5.56 & 4.76 & 2.90 & 2.46 & 1.95 & \textbf{1.87} \\
 & D3I & 17.24 & 22.68 & 10.02 & \textbf{6.73} & 5.90 & \textbf{3.45} & 3.81 & 3.58 & 2.17 & 1.70 & 1.64 & \textbf{1.46} \\ \bottomrule
\end{tabular}
\label{tab:combined_nomad_wno}
\caption*{\textit{Note: Ind. = Individual, MH = Multi-Head, SH-B = Single-Head Base, SH-F = Single-Head Full. SPH = SPHER, KST = KSTAR, D3I = D3D-ITER.}}
\end{table}

\begin{table}[h!]
\centering
\captionsetup{justification=centering}
\caption{Relative $L2$ Error (\%) for Unconstrained Transfer Learning Strategies across 10, 100, and 1000 Samples for Sp$^2$GNO and WNO Models}
\begin{tabular}{@{}llcccccccccccc@{}}
\toprule
\multirow{2}{*}{\textbf{Model}} & \multirow{2}{*}{\textbf{Task}} & \multicolumn{4}{c}{\textbf{10 Samples}} & \multicolumn{4}{c}{\textbf{100 Samples}} & \multicolumn{4}{c}{\textbf{1000 Samples}} \\ \cmidrule(lr){3-6} \cmidrule(lr){7-10} \cmidrule(lr){11-14}
 &  & \textbf{Ind.} & \textbf{MH} & \textbf{SH-B} & \textbf{SH-F} & \textbf{Ind.} & \textbf{MH} & \textbf{SH-B} & \textbf{SH-F} & \textbf{Ind.} & \textbf{MH} & \textbf{SH-B} & \textbf{SH-F} \\ \midrule
\multirow{3}{*}{\rotatebox[origin=c]{90}{\tiny Sp$^2$GNO}} & SPH & 53.27 & \textbf{29.38} & 49.45 & 48.30 & 26.23 & \textbf{10.70} & 31.85 & 32.10 & 10.91 & \textbf{3.73} & 12.61 & 12.21 \\
 & KST & 63.53 & \textbf{42.45} & 56.79 & 48.73 & 25.90 & \textbf{12.56} & 19.05 & 19.70 & 8.77 & \textbf{3.87} & 11.05 & 10.71 \\
 & D3I & 46.70 & \textbf{20.06} & 35.20 & 33.52 & 24.17 & \textbf{9.83} & 20.54 & 19.34 & 8.88 & \textbf{4.37} & 10.74 & 10.37 \\ \cmidrule(l){2-14} 
\multirow{3}{*}{\rotatebox[origin=c]{90}{\tiny Sp$^2$GNO-Skip}} & SPH & 23.98 & \textbf{16.27} & 36.69 & 28.02 & 4.67 & \textbf{3.21} & 4.77 & 6.94 & 0.80 & \textbf{0.45} & 1.08 & 1.03 \\
 & KST & 62.37 & 46.46 & 65.94 & \textbf{28.64} & 14.05 & \textbf{4.81} & 20.60 & 7.89 & 1.32 & \textbf{0.91} & 1.66 & 1.30 \\
 & D3I & 56.09 & \textbf{21.30} & 75.46 & 27.08 & 10.19 & \textbf{3.43} & 10.55 & 5.86 & 0.98 & \textbf{0.72} & 1.37 & 1.08 \\ \cmidrule(l){2-14} 
\multirow{3}{*}{\rotatebox[origin=c]{90}{\tiny WNO}} & SPH & 14.01 & 10.96 & 10.17 & \textbf{8.92} & 6.26 & 3.89 & 3.36 & \textbf{3.07} & 1.62 & 1.52 & \textbf{1.07} & 1.28 \\
 & KST & 18.57 & \textbf{13.52} & 19.20 & 13.55 & 6.89 & \textbf{4.64} & 5.56 & 4.76 & 2.90 & 2.46 & 1.95 & \textbf{1.87} \\
 & D3I & 17.24 & 22.68 & 10.02 & \textbf{6.73} & 5.90 & \textbf{3.45} & 3.81 & 3.58 & 2.17 & 1.70 & 1.64 & \textbf{1.46} \\ \bottomrule
\end{tabular}
\label{tab:skipped_sp2_wno}
\caption*{\textit{Note: Ind. = Individual, MH = Multi-Head, SH-B = Single-Head Base, SH-F = Single-Head Full. SPH = SPHER, KST = KSTAR, D3I = D3D-ITER.}}
\end{table}

\section{Memory and Power Efficiency}

Tables~\ref{tab:model-size-latency-pretrain} and
\ref{tab:model_inference_power} report the GPU memory footprint and
instantaneous power draw for all five architectures across single-task
training, multi-task training, transfer fine-tuning, and inference,
evaluated on a single NVIDIA A100 GPU. No implementation-level memory
or compute optimizations were applied; figures therefore reflect the
raw requirements of each architecture and should be interpreted
relative to one another rather than as absolute lower bounds.

\paragraph{Memory footprint.}
FNO requires the least memory for single-task training at $1.59$~GiB
despite having the largest parameter count ($2{,}180{,}737$), which
reflects the efficiency of its structured Fourier convolution relative
to the dense MLP computations in branch-trunk and graph-based models.
MIONet, NOMAD, and WNO occupy a similar range of $2.13$--$3.11$~GiB
under single-task training, broadly consistent with their parameter
counts and computational graphs. Multi-task training increases memory
for all architectures, with the largest relative growth observed for
NOMAD ($2.39 \to 3.44$~GiB, $44\%$ increase) and the smallest for
FNO ($1.59 \to 2.14$~GiB, $35\%$ increase).

Sp$^2$GNO stands apart from all other architectures in its memory scaling. Under single-task training, it already requires $10.50$~GiB, 6.6 times the memory of FNO and 3.4 times that of WNO. Under multi-task training, this rises to $35.31$~GiB, 10.3 times the memory of the next highest architecture (NOMAD at $3.44$~GiB) and 9.4 times that of WNO ($3.77$~GiB). This disproportionate scaling arises from constructing and storing the graph Laplacian eigenmodes for all five source-task meshes simultaneously, a cost that grows with the number of tasks and the density of the graph.
Transfer fine-tuning reduces Sp$^2$GNO's memory to $5.64$~GiB, an $84\%$ reduction relative to multi-task training, since freezing the backbone eliminates the activations and intermediate graph representations required for full backpropagation. Its inference memory of $5.64$~GiB
reflects the cost of maintaining the graph structure at evaluation time. All other architectures maintain inference memory within $1.50$--$3.16$~GiB, with FNO achieving the lowest inference footprint at $1.50$~GiB and WNO the highest among the non-graph models at $3.04$~GiB.

Multi-head variants add marginal parameter overhead across all
architectures (e.g., $66{,}564$ additional parameters for FNO),
consistent with the lightweight nature of the task-specific projection
layers. The corresponding memory increase for multi-head inference is
similarly small ($\leq 0.13$~GiB for FNO, MIONet, NOMAD, and WNO),
with the exception of Sp$^2$GNO, where multi-head inference rises to
$9.07$~GiB due to the parallel evaluation of five graph-based
projection pathways.

\paragraph{Power consumption.}
WNO records the lowest training power draw of any architecture at
$86.73$~W for single-task training, rising only marginally to
$89.70$~W ($3.4\%$ increase) under multi-task training. This near-
constant power profile indicates that WNO's wavelet convolutions are
computed at consistently low GPU utilization regardless of the number
of concurrent tasks. MIONet is the second most power-efficient
architecture during training ($93.92$~W single-task) and achieves the
lowest inference power of all models at $81.79$~W, making it the most
energy-efficient option for repeated inference at deployment.

FNO and NOMAD occupy an intermediate power band, with single-task
training draws of $134.78$~W and $146.23$~W, respectively, rising to
$182.27$~W and $185.27$~W under multi-task training. NOMAD's
multi-head inference power of $172.61$~W is notably higher than its
single-head inference draw of $126.36$~W, reflecting the cost of
evaluating five separate decoder MLPs in parallel.

Sp$^2$GNO draws the highest power across all operational modes, consuming $205.56$~W during single-task training and $228.07$~W during multi-task training, 2.54 times and 2.54 times the power of WNO in the respective modes. Its inference power of $208.31$~W is 2.55 times that of MIONet ($81.79$~W) and exceeds FNO's inference draw ($128.25$~W) by $63\%$. Unlike its memory footprint, Sp$^2$GNO's power consumption does not scale as dramatically between single-task and multi-task training ($205.56 \to 228.07$~W, an $11\%$ increase), suggesting that the multi-task memory overhead is driven by stored graph representations rather than increased active computation.

\paragraph{Summary.}
Across both metrics, FNO and MIONet are the most resource-efficient
architectures: FNO minimizes training and inference memory, while
MIONet minimizes inference power. WNO achieves the lowest training
power and scales its power draw minimally under multi-task training,
making it the most energy-efficient option for the pre-training phase.
Sp$^2$GNO incurs the highest memory and power costs at every stage,
with multi-task training memory an order of magnitude above all other
architectures; this overhead is substantially recovered during transfer
fine-tuning through backbone freezing. These efficiency profiles
complement the accuracy analysis in the main text: WNO's low training
power and moderate memory footprint, combined with its leading transfer
accuracy, make it the most favorable architecture overall for the
pre-train and fine-tune deployment scenario considered in this work.

\begin{table}[h!]
\centering
\caption{Model complexity and average memory required (GB) over 10 epochs. For Single Tasks, a single model and task were chosen. All memory data is evaluated on an Nvidia A100 GPU. It should be noted that optimization over code implementation was not performed. Analysis should be focused on model requirements relative to each other.}
\begin{tabular}{@{}ccccccc@{}}
\toprule
\multirow{2}{*}{\textbf{Model}} & \multirow{2}{*}{\textbf{MLP-Head Count}} & \multirow{2}{*}{\textbf{Parameters}} & \multicolumn{3}{c}{\textbf{Training Memory (GiB)}} & \multirow{2}{*}{\textbf{Inf. Memory (GiB)}}    \\ \cmidrule(l){4-6} 
& & & Single Task & Multi-Tasks & Transfer & \\ \midrule
\multirow{2}{*}{\textbf{FNO}}  & Single Head & 2,180,737 & 1.59 & 2.14 & 1.53 & 1.50\\
 & Multi-Head & 2,247,301 & - & 2.16 & - & 1.57\\
\cmidrule(l){2-7}
\multirow{2}{*}{\textbf{MIONet}}  & Single Head & 249,088 & 2.13 & 2.49 & 2.10 & 2.05 \\
 & Multi-Head & 579,328 & - & 2.49 & - & 2.15\\
\cmidrule(l){2-7}
\multirow{2}{*}{\textbf{NOMAD}}  & Single Head & 364,289 & 2.39 & 3.44 & 2.26 & 2.26 \\
 & Multi-Head & 430,953 & - & 3.44 & - & 2.60\\
\cmidrule(l){2-7}
\multirow{2}{*}{\textbf{Sp$^2$GNO}}  & Single Head & 1,497,053 & 10.50 & 35.31 & 5.64 & 5.64 \\
 & Multi-Head & 1,563,617 & - & 35.31 & - & 9.07\\
 \cmidrule(l){2-7}
\multirow{2}{*}{\textbf{WNO}}  & Single Head & 1,787,521 & 3.11 & 3.77 & 3.04 & 3.04 \\
 & Multi-Head & 1,854,085 & - & 3.75 & - & 3.16\\
\bottomrule
\end{tabular}
\label{tab:model-size-latency-pretrain}
\end{table}

\begin{table}[h!]
\centering
\caption{Model complexity and average instantaneous power (W) over 10 epochs. For Single Tasks, a single model and task were chosen. All memory data is evaluated on an Nvidia A100 GPU. It should be noted that optimization over code implementation was not performed. Analysis should be focused on model requirements relative to each other.}
\begin{tabular}{@{}ccccccc@{}}
\toprule
\multirow{2}{*}{\textbf{Model}} & \multirow{2}{*}{\textbf{MLP-Head Count}} & \multirow{2}{*}{\textbf{Parameters}} & \multicolumn{3}{c}{\textbf{Training Power (W)}} & \multirow{2}{*}{\textbf{Inf. Power (W)}}    \\ \cmidrule(l){4-6} 
& & & Single Task & Multi-Tasks & Transfer & \\ \midrule
\multirow{2}{*}{\textbf{FNO}}  & Single Head & 2,180,737 & 134.78 & 182.27 & 133.35 & 128.25\\
 & Multi-Head & 2,247,301 & - & 178.56 & - & 164.08\\
\cmidrule(l){2-7}
\multirow{2}{*}{\textbf{MIONet}}  & Single Head & 249,088 & 93.92 & 133.52 & 106.21 & 81.79 \\
 & Multi-Head & 579,328 & - & 121.50 & - & 122.20\\
\cmidrule(l){2-7}
\multirow{2}{*}{\textbf{NOMAD}}  & Single Head & 364,289 & 146.23 & 185.27 & 132.24 & 126.36 \\
 & Multi-Head & 430,953 & - & 194.14 & - & 172.61\\
\cmidrule(l){2-7}
\multirow{2}{*}{\textbf{Sp$^2$GNO}}  & Single Head & 1,497,053 & 205.56 & 228.07 & 203.47 & 208.31 \\
 & Multi-Head & 1,563,617 & - & 226.95 & - & 207.21\\
 \cmidrule(l){2-7}
\multirow{2}{*}{\textbf{WNO}}  & Single Head & 1,787,521 & 86.73 & 89.70 & 101.98 & 101.07 \\
 & Multi-Head & 1,854,085 & - & 89.83 & - & 106.98\\
\bottomrule
\end{tabular}
\label{tab:model_inference_power}
\end{table}

\section{Training Details}
Table~\ref{tab:model-architectures} lists the architecture
specifications for all five neural operator models. For the
branch-trunk architectures (MIONet and NOMAD), each scalar input
($A$, $\varepsilon$, $\kappa$, $\delta$) is assigned an independent
branch network with input dimension 1, and the spatial coordinates
$(x, y)$ are processed by a trunk network with input dimension 2.
All branch and trunk MLPs use a hidden and output width of 128 across
four layers. NOMAD augments this structure with an additional combined
network of architecture $[640, 128, 128, 128, 1]$, where the input
dimension of 640 corresponds to the concatenation of the four
branch output vectors ($4 \times 128$) and the trunk output vector
($128$). For FNO, WNO, and Sp$^2$GNO, the input at each evaluation
point is the concatenation of all four scalar parameters with the
local coordinates, processed through four spectral or graph operator
blocks of width 128. The WNO uses a level-4 wavelet decomposition
with $\omega = 8$, the FNO retains 4 Fourier modes per layer, and
Sp$^2$GNO constructs a $k = 50$ KNN graph with 10 spectral modes.
All downlift projection MLPs use GeLU activation; branch and trunk
MLPs use ReLU; and the post-block nonlinearities for FNO, WNO, and
Sp$^2$GNO are ReLU, Mish, and GeLU, respectively.

Table~\ref{tab:training_hyperparameters} lists the training
hyperparameters applied uniformly across all architectures for both
full-training and transfer learning. A batch size of 32 is used in
both regimes. Full-training runs for a maximum of 150 epochs with an
initial learning rate of $10^{-3}$, while transfer learning runs for
a maximum of 100 epochs with a higher initial learning rate of
$10^{-2}$, which accelerates convergence when only the lightweight
projection head is being updated. Both regimes apply constant step
decay with a decay rate of 0.5 every 10 epochs and use the Adam
optimizer. Weight decay is set to $10^{-5}$ during full-training and
reduced to $10^{-6}$ during transfer to limit regularization of the
small number of trainable parameters. Early stopping with a patience
of 50 epochs is applied in both regimes to prevent overfitting and
reduce unnecessary computation when validation loss plateaus.

\begin{table}[h!]
\centering
\caption{Architecture specifications for baseline and proposed models. All trunk-branch networks use ReLU activation, downlift MLP's utilize GeLU, and FNO, WNO, and Sp$^2$GNO use ReLU, Mish, and GeLU respectively after their spectral convolution blocks. The input dimension for each branch net is 1, corresponding to the input parameters: $A$, $\varepsilon$, $\kappa$, and $\delta$.}
\label{tab:model-architectures}
\begin{tabular}{@{}llc@{}}
\toprule
\textbf{Model Name} & \textbf{Branch Net Architecture} & \textbf{Trunk Net Architecture} \\ \midrule
MIONet & $[1, 128, 128, 128, 128]$ & $[2, 128, 128, 128, 128]$ \\
NOMAD & $[1, 128, 128, 128, 128]$ & $[2, 128, 128, 128, 128]^*$ \\
FNO & \multicolumn{2}{c}{4 Fourier layers with width 128 and 4 modes} \\
WNO & \multicolumn{2}{c}{4 Wavelet layers with width 128 and level 4 decomp and omega = 8} \\
Sp$^2$GNO & \multicolumn{2}{c}{4 Block layers with width 128, 10 modes, and $k=50$ KNN graph}
\\ \bottomrule
\multicolumn{3}{l}{\footnotesize $^*$NOMAD includes an additional combined network of $[640, 128, 128, 128, 1]$.}
\end{tabular}
\end{table}

\begin{table}[h!]
\caption{Training Hyperparameters}
  \centering
\begin{tabular}{@{}ccc@{}}
\toprule
\textbf{Parameter} & \textbf{Full-Training} & \textbf{Transfer Learning}
\\ \midrule
\textbf{Batch Size}                  & 32                              & 32         \\
\textbf{Max Epochs}                & 150                               & 100    \\
\textbf{Learning Rate}                    & 1e-3                              & 1e-2    \\
\textbf{Decay Steps}                    & 10                           & 10   \\
\textbf{Decay Rate}                    & 0.5                           & 0.5      \\
\textbf{Optimizer}                    & Adam                           & Adam     \\
\textbf{Weight Decay}                    & 1e-5                           & 1e-6     \\
\textbf{Early Patience}                    & 50                           & 50      \\
\bottomrule
\end{tabular}
\label{tab:training_hyperparameters}
\end{table}

\end{document}